\documentclass[preprint,12pt]{elsarticle}

\usepackage{lineno,hyperref,amsmath}
\usepackage{geometry}
\usepackage{amssymb,mathrsfs}
\usepackage{amsthm}
\usepackage{booktabs}
\usepackage{graphicx}
\usepackage{subfigure}
\usepackage[]{caption2}
\usepackage{enumerate}
\usepackage{color}
\usepackage{bm}
\usepackage{picinpar,wrapfig}
\usepackage{setspace}
\usepackage{algorithm}
\usepackage{algpseudocode}
\usepackage{mathrsfs}
\usepackage{rotating}
\biboptions{numbers,sort&compress}

\newtheorem{myDef}{Definition}[section]
\journal{Neurocomputing}

\begin{document}
\begin{frontmatter}



\title{Rough extreme learning machine: a new classification method based on uncertainty measure \tnoteref{mytitlenote}}


\author[address1]{Lin Feng \corref{mycorrespondingauthor}}
\cortext[mycorrespondingauthor]{Corresponding author}
\ead{fenglin@dlut.edu.cn}
\author[address2]{Shuliang Xu}
\ead{xushulianghao@126.com}
\author[address1]{Feilong Wang}
\author[address1]{Shenglan Liu}

\address[address1]{School of Innovation and Entrepreneurship, Dalian University of Technology, Dalian, China}
\address[address2]{Faculty of Electronic Information and Electrical Engineering, Dalian University of Technology, Dalian , China}

\begin{abstract}
Extreme learning machine (ELM) is a new single hidden layer feedback neural network. The weights of the input layer and the biases of neurons in hidden layer are randomly generated; the weights of the output layer can be analytically determined. ELM has been achieved good results for a large number of classification tasks. In this paper, a new extreme learning machine called rough extreme learning machine (RELM) was  proposed. RELM uses rough set to divide data into upper approximation set and lower approximation set, and the two approximation sets are utilized to train upper approximation neurons and lower approximation neurons. In addition, an attribute reduction is executed in this algorithm to remove redundant attributes. The experimental results showed, comparing with the comparison algorithms, RELM can get a better accuracy and repeatability in most cases; RELM can not only maintain the advantages of fast speed, but also effectively cope with the classification task for high-dimensional data.
\end{abstract}


\end{frontmatter}


\section{Introduction}
 With the development of information society, various industries have produced a great deal of data and how to effectively analyze these data becomes an urgent problem \cite{Wu2013Data}. Classification is as a basic form of data analysis; it has attracted scholars much attention \cite{Han2001Data,Parker2003Empirical}. In 2006, extreme learning machine is proposed by Huang as a new classification method\cite{Huang2006Extreme,Huang2012Extreme,Huang2010Optimization,Huang2008Enhanced}. In recent years, ELM has been extensively studied by researchers \cite{Zhang2016Incremental,Tang2017Extreme,Cornejo2016A}. Zhang et al. proposed a privileged knowledge extreme learning machine called ELM+ for radar signal recognition \cite{Zhang2015A}; in practical applications, many classification tasks has privileged knowledge, but the traditional ELM \cite{Huang2006Extreme,Huang2012Extreme} does not take advantage of these privileged knowledge; ELM+ makes full use of privileged knowledge to map input data into a feature space and a correction space; it uses the traditional ELM and the privileged knowledge to get a corrected hidden layer output matrix and output layer weights; by solving the corrected optimization problem, the solution of the ELM+ is obtained. Aiming at classification in blind domain adaptation, Uzair et al. developed a new ELM model named A-ELM \cite{Uzair2017Blind}. In order to cope with the problem that there is a big difference between the distribution of training data and target domain data, A-ELM uses a multiple classifiers system; For A-ELM algorithm, a global classifier is trained by whole training data which can classify all classes data; then the data is divided into \emph{c} subsets (\emph{c} is the number of classes) and \emph{c} local classifiers are also trained based on the \emph{c} subsets. When a new data coming, the algorithm utilizes the (\emph{c}+1) classifiers to classify the new data. The local classier whose square error is least with global classifier is as final training classification model. Deng et al. proposed a fast and accurate kernel-based supervised extreme learning machine referred to as RKELM \cite{Deng201629}. RKELM introduces kernel function and random selection mechanism to improve the performance of the algorithm. Aiming at reducing the size of the output matrix of the hidden layer, support vectors of RKELM is are randomly selected from training set and the number of support vectors is limited to less than the number of neurons in hidden layer. The traditional ELM is easily affected by noise, but with fast speed; sparse representation classification (SRC) has a good ability to resist noise, but its speed does not have too many advantages. So Cao et al. designed a new extreme learning machine method based on adaptive sparse representation for image classification called EA-SRC \cite{Cao2016Extreme}. EA-SRC combines ELM with SRC, and employs regularization mechanism to improve generalization performance. For the sake of optimal regularization parameter selection, it adopts the leave-one-out cross validation (LOO) scheme. In addition, considering reducing computational complexity in ${H}^{T}H$ or $H{H}^{T}$, EA-SRC uses SVD(Singular Value Decomposition) to reduce the dimensions of ${H}^{T}H$ or $H{H}^{T}$. Li et al. proposed an extreme learning machine method with transfer learning mechanism called TL-ELM \cite{Li2016Extreme}. Different from the traditional ELM, TL-ELM asks the difference between the old domain knowledge and the new domain knowledge must be as small as possible. By solving the optimization problem, the output weights of the new ELM can be got. For imbalanced and big data classification problem, Wang et al. proposed a distributed weighted extreme learning machine  referred to as DWELM \cite{Wang2016}. To handle big data and imbalanced data, DWELM draws into MapReduce framework and sample weighting method. Aiming at data stream classification containing concept drift, Mirza et al. designed a meta-cognitive online sequential extreme learning machine called MOS-ELM \cite{Mirza2016Meta}. MOS-ELM is a development of OS-ELM \cite{Huang2005On}, and uses online sequential learning method to learn data stream and deal with concept drift \cite{Bifet2014A,Xu2016A,Lu2014Concept,Xu2017Dynamic}; the sliding window can be adaptively adjusted according to the accuracy of classification; if a sample is correctly classified, it will be deleted from sliding window; if misclassified, it will be added to sliding window again and then SMOTE algorithm \cite{Chawla2002SMOTE} is executed to retrain classifier.

 Rough set is a mathematical tool to analyze data proposed by Z. Pawlak \cite{Pawlak1982Rough}. Because it can deal with imprecise, inconsistent and incomplete information and eliminate redundant attributes from feature sets without any preliminary or prior information, it has been widely used in pattern recognition, image processing, biological data analysis, expert systems and other fields in recent years \cite{An2016Data,Meng2015Gene,Kim2016Developing}. Some researchers have investigated methods to combine rough set with neural network to better get classification models \cite{Lingras1998Comparison}. Kothari et al. applied rough set theory in the architecture of unsupervised neural network \cite{Kothari2008Rough} and the proposed algorithm uses the Kohonen learning rule to train neural network. Azadeh proposed an integrated data envelopment analysis$\textrm{-}$artificial (DEA) neural network$\textrm{-}$rough set algorithm for assessment of personnel efficiency \cite{Azadeh2011An}; at first, it uses rough set to determine many candidate reductions and then calculates the performance of neural network for each reduct; the best reduct is selected by ANN results though DEA. Ahn et al. proposed a hybrid intelligent system combining rough set with artificial neural network to predict the failure of firms \cite{Ahn2000The} and a new reduction algorithm called 2D reduction was designed; in 2D reduction method, rough set is utilized to eliminate irrelevant or redundant attributes and then scans samples to delete the samples of inconsistent decisions; at last, association rules can be extracted from the data; for Ahn's hybrid classification model, if a new instance matches some association rules, the instance will be classified by the association rules; otherwise, the algorithm will use the data reduced by 2D reduction method to train a classifier to classify this instance. Xu et al. introduced a rough rule granular extreme learning machine called RRGELM\cite{Xu2015A}; RRGELM uses rough set to extract association rules and the number of neurons in hidden layer is decided by the number of the association rules; the weights of input layer is not randomly generated, they are determined by the conditions whether instances are covered by the association rules or not. The above works have promoted the developments of rough set and neural network. However, those models only utilize rough set to reduce attributes or the cost of training rough neural networks is too large. Inspired by the above models, a new classification method combined extreme learning machine with rough set referred to as RELM was proposed in this paper. For RELM, the input weights and biases of hidden layer are randomly generated, and the training set is divided into two parts: upper approximation set and lower approximation set to train upper approximation neurons and lower approximation neurons. In addition, attributes reduction is introduced to eliminate the influence of redundant attributes on classification results. Because the input weights and biases of hidden layer are randomly generated and the output weights can analytically determined, RELM can overcome the some disadvantages of conventional neural networks and has a fast training speed.

 The contributions of this paper are as follows:
 \begin{itemize}
   \item A new extreme learning machine is designed in this paper. Different from the traditional ELM \cite{Huang2006Extreme}, RELM utilizes rough set to divide data into upper approximation set and lower approximation set and then uses the upper approximation set and lower approximation set to train upper boundary neurons and lower boundary neurons correspondingly. Every neuron of RELM cantains two neurons: a upper boundary neuron and a lower boundary neuron. The final classification result is decided by the two kind of neurons.
   \item Attribute reduction is introduced for RELM. Rough set has obvious advantages in attribute reduction and it can preprocess data according to the data itself and dose not need any prior knowledge. By using rough set, RELM can remove redundant attributes without any information loss and improve the performance of the proposed extreme learning machine algorithm.
   \item A new method for determining the number of neurons in hidden layer is proposed. In RELM, the number of neurons in hidden layer is determined by the sizes of positive region and boundary region. It can reduce the blindness of selecting the number of neurons in hidden layer.
   \item Rough set is used to guide the learning process of ELM. Traditional algorithms often separate rough sets and neural networks and do not fuse ELM and rough sets very well. Different from existing algorithms which only use rough set to reduce attributes or determine the number of neurons in hidden layer \cite{Ahn2000The,Xu2015A}, RELM uses the results of the data divided by rough set to train different kinds of neurons; it preferably combines rough set with extreme learning machine.
 \end{itemize}

 The rest of this paper is organized as follows. In Section 2, the basic concepts and principles of ELM and rough set are reviewed. Section 3 introduces the proposed algorithm and describes the implementation process and principle of RELM. In Section 4, RELM and comparison algorithms are evaluated on data sets and the results are analysed in detail. Finally, the conclusions are stated in section 5.

\section{Preliminaries}
In this section, we give a description of necessary preparatory knowledge in this paper. Firstly, we look into the essence of extreme learning machine and introduces the steps of ELM. Then rough set is reviewed and we describes the basic concepts and principles of rough set in detail.
\subsection{The model of ELM}
ELM is a single-hidden layer feedback neural network; the input weights and biases of hidden layer nodes are randomly selected, and the output weights can be analytically determined by the least square method. ELM is very fast in speed and has a good generalization ability \cite{Tang2017Extreme}.

For \emph{N} distinct sample $\left \{ (\bm{x}_{i},\bm{t}_{i})_{i=1}^{N} \right \}$, $\bm{x}_{i}=\left \{ x_{i1},x_{i2},\cdots ,x_{im} \right \}^{T}\in \mathbb{R}^{m}$ and $\bm{t}_{i}=\left \{ t_{i1},t_{i2},\cdots ,t_{in} \right \}^{T}\in \mathbb{R}^{n}$, the output of ELM with \emph{L} hidden layer nodes is as \cite{Feng2009Error,Sun2017Efficient}:
\begin{equation}
  f(\bm{x}_{i})=\sum_{j=1}^{L}\bm{\beta} _{j}g(\bm{w}_{j},b_{j},\bm{x}_{j})  \ \ \ \ \ \ \ i=1,2,\cdots ,N
\end{equation}
where $\bm{w}_{j}=\left \{ w_{j1},w_{j2},\cdots ,w_{jm} \right \}$ is the input weights between the nodes of input layer and the $\emph{j}_{th}$ nodes of hidden layer; $\bm{\beta}_{j}$ is the output weights of the $\emph{j}_{th}$ nodes of hidden layer and $g(\cdot)$ is the active function of hidden nodes which is a nonlinear piecewise continuous function. From the literatures \cite{Huang2006Extreme,Wang2016Self}, ELM can approximate any target function with zero error. So Eq.(1) can be written as:
\begin{equation}
  \bm{H}\bm{\beta}=\bm{T}
\end{equation}
$\bm{H}$ is output matrix of hidden layer nodes:
\begin{equation}
  H=\begin{bmatrix}
h(\bm{x}_{1})\\
\vdots \\
h(\bm{x}_{N})
\end{bmatrix}=\begin{bmatrix}
g(\bm{w}_{1},b_{1},\bm{x}_{1}) &\cdots   &g(\bm{w}_{L},b_{L},\bm{x}_{1}) \\
 \vdots & \cdots  & \vdots \\
 g(\bm{w}_{1},b_{1},\bm{x}_{N})& \cdots  &g(\bm{w}_{L},b_{L},\bm{x}_{N})
\end{bmatrix}_{N\times L}
\end{equation}
$\bm{\beta}$ is the output weights of hidden nodes and $\bm{T}$ is the target matrix of ELM, where
\begin{equation}
  \bm{\beta}=\begin{bmatrix}
\bm{\beta}_{1}^{T}\\
\vdots \\
\bm{\beta}_{L}^{T}
\end{bmatrix}_{L\times m} \ \ \ \ \ \text{and} \ \ \ \ \ \bm{T}=\begin{bmatrix}
\bm{t}_{1}^{T}\\
\vdots \\
\bm{t}_{N}^{T}
\end{bmatrix}_{N\times m}
\end{equation}
The smallest norm least-squares solution of Eq.(2) \cite{Huang2008Incremental} is
\begin{equation}
  \bm{\beta}=H^{\dagger }\bm{T}
\end{equation}
$\bm{H}^{\dagger}$ is the Moore-Penrose generalized inverse \cite{Xu2017Dynamic}, and it can be calculated by orthogonal projection method, orthogonalization method and singular value composition (SVD) \cite{Zhangxianda}.

In order to reduce the influence of ill conditioned matrix on the calculation results and improve the robustness, ridge parameter is used in ELM \cite{Cambria2013Extreme,Huang2011Extreme,Huang2015Trends}. The optimization problem of ELM can be described as
\begin{equation}
\begin{split}
  & Min: \ \frac{1}{2}\left \| \bm{\beta}  \right \|^{2}+\frac{C}{2}\sum_{i=1}^{N}\left \| \bm{\xi} _{i} \right \|^{2}\\
  & s.t.\ \ \ \ \ \ \bm{h}(\bm{x}_{i})\bm{\beta }=\bm{t}_{i}^{T}-\bm{\xi}_{i}^{T}, \ \ i=1,2,\cdots ,N
\end{split}
\end{equation}
Where \emph{C} is penalty factor; $\bm{\xi}_{i} $ is the resident between target value and real value of the $i_{th}$ sample. According to KKT conditions, if $L>N$, the solution of Eq.(6) can be expressed as \cite{Huang2015Trends}:
\begin{equation}
  \bm{\beta}=H^{\dagger }\bm{T}=({\frac{\bm{I}}{C}+\bm{H}^{T}\bm{H}})^{-1}\bm{H}^{T}\bm{T}
\end{equation}
So the output of ELM is
\begin{equation}
  f(\bm{x})= h(\bm{x})\bm{\beta}=h(\bm{x})({\frac{\bm{I}}{C}+\bm{H}^{T}\bm{H}})^{-1}\bm{H}^{T}\bm{T}
\end{equation}
If $L\leq N$, the solution of Eq.(6) can be expressed as
\begin{equation}
  \bm{\beta}=H^{\dagger }\bm{T}=\bm{H}^{T}({\frac{\bm{I}}{C}+\bm{H}\bm{H}^{T}})^{-1}\bm{T}
\end{equation}
So the output of ELM is
\begin{equation}
  f(\bm{x})= h(\bm{x})\bm{\beta}=h(\bm{x})\bm{H}^{T}({\frac{\bm{I}}{C}+\bm{H}\bm{H}^{T}})^{-1}\bm{T}
\end{equation}
For binary classification problem, the decision result of ELM is:
\begin{equation}
  label(\bm{x})=\text{sign}(h(\bm{x})\bm{\beta})
\end{equation}
For classification problem, the decision result of ELM is
\begin{equation}
  label(\bm{x})=\mathop{\text{argmax}}\limits_{i\in \left \{ 1,2,\cdots ,n \right \}}f_{i}(\bm{x})
\end{equation}
Where $f(\bm{x})=\left [ f_{1}(\bm{x}),f_{2}(\bm{x}),\cdots ,f_{n}(\bm{x}) \right ]$.

From the above descriptions, the main steps of ELM are summarized as follows:
 \begin{algorithm} \caption{ ELM model.}
  \begin{algorithmic}
   \Require
    a training data set $\mathcal{X}=\left \{ (\bm{x}_{i},\bm{y}_{i})|\bm{x}_{i} \in \mathbb{R}^{m}, \bm{y}_{i} \in \mathbb{R}^{n} \right \}$;
    the number of nodes in hidden layer \emph{L}.
    the activation function $g(\cdot)$;
    \Ensure ELM classifier.
    \State $\bm{Step \ 1}$: Randomly generate the input weights $\bm{w}_{j}$ and biases $\bm{b}_{j}$, $j=1, 2,\cdots, L$;
    \State $\bm{Step \ 2}$: Calculate the output matrix of hidden layer $\bm{H}$ for data set $\mathcal{X}$;
    \State $\bm{Step \ 3}$: Obtain the output weights $\bm{\beta}$ according to Eq.(7) or Eq.(9);
   \end{algorithmic}
\end{algorithm}

\subsection{The basics of rough set theory}
Rough set is a useful tool for the study of imprecise and uncertain knowledge. Rough set can analyze categorical data according to data itself and it dose not need any prior knowledge; it has been successfully applied in attribute reduction \cite{Chen2015A}. Rough set claims that the objects with the same attribute value should belong to the same decision class, otherwise, it violates the principle of classification consistency \cite{Pawlakab1997Rough}.

Let $IS=\left \langle U,A,V,f \right \rangle$ be an information system \cite{Shu2014Incremental}, where $U=\left \{ x_{1},x_{2},\cdots ,x_{n} \right \}$ is the universe with a non-empty set of finite objects, $A=\left \{ a_{1},a_{2},\cdots ,a_{m} \right \}$ is an attributes set, $V_{a}$ is the values set when the attribute is \emph{a} and $\emph{V}=\sum\limits_{a\in A}V_{a}$ is the values set of attributes set \emph{A}. \emph{f} is a information function: if $\forall x\in U, \forall a\in A$, it has $f(x,a)\in V_{a}$. For an information system, $A=C\cup D$ and $C\cap D=\varnothing $, where \emph{C} is a conditional attributes set and \emph{D} is a decision attribute set; the information system is also be named decision table. So it has the following definitions from the literatures \cite{Lazo2015On,YAO201640,Luan2016A}.

\begin{myDef}
\normalfont{For an information system $IS=\left \langle U,A,V,f \right \rangle$, let $B\subseteq A$ and the relation between attributes set \emph{B} and universe is defined as:
\begin{equation}
  IND(B)= \left \{ (x,y)\in U\times U|f(x,a)=f(y,a),\forall a\in B \right \}
\end{equation}
\emph{IND(B)} is called indiscernibility relation (equivalence relation).
}
\end{myDef}

It is known that the objects \emph{x} and \emph{y} have the same values on the attribute set \emph{B} if $(x,y)\in IND(B)$ and it can also say \emph{x} and \emph{y} are indiscernible by attributes set \emph{B}.
\begin{myDef}
\normalfont{ For $\forall x\in U $, the \emph{x}'s equivalence class of the \emph{B}-indiscernibillity relation is defined as follow:
\begin{equation}
  \left [  x\right ]_{B}=\left \{ y\in U|(x,y)\in IND(B) \right \}
\end{equation}
The partition of \emph{U} generated by \emph{IND(B)} is denoted \emph{U/IND(B)} or abbreviated as \emph{U/B}.
}
\end{myDef}

From Definition 2.2, it is known that equivalence class is a set that all objects has the same attributes values; if $U/B=\left \{ U_{1},U_{2},\cdots,U_{n}\right \}$, it has $U=U_{1}\cup U_{2}\cup \cdots \cup U_{n}$ and $U_{i}\cap U_{j}=\varnothing$ for $\forall i\neq j$.

\begin{myDef}
\normalfont{Let $IS=\left \langle  U,A,V,f\right \rangle $ be an information system, and $R_{B}$ is an equivalence relation of universe \emph{U} generated by $B\subseteq A$; we call $\left \langle U, R_{B} \right \rangle $ is an approximation space, If $X\subseteq U$, so the $R_{B}$-lower approximation $\underline{R_{B}}X$ and $R_{B}$-upper approximation $\overline{R_{B}}X$ are defined as:
\begin{equation}
  \underline{R_{B}}X=\left \{ x\in U|\left [  x\right ]_{B}\subseteq X \right \} \ \ \ \ \text{and} \ \ \ \
  \overline{R_{B}}X=\left \{ x\in U|\left [  x\right ]_{B}\cap X \neq \varnothing \right \}
\end{equation}
}
\end{myDef}
Rough set divides the universe into three regions: positive region, negative region and boundary region; it can be seen as Fig.1.

\emph{B}-positive region of \emph{X} is as:
\begin{equation}
  POS_{B}(X)=\underline{R_{B}}X
\end{equation}

\emph{B}-negative region of \emph{X} is as:
\begin{equation}
  NEG_{B}(X)=U-\overline{R_{B}}X
\end{equation}

\emph{B}-boundary region of \emph{X} is as:
\begin{equation}
  BN_{B}(X)=\overline{R_{B}}X-\underline{R_{B}}X
\end{equation}
\begin{figure}
  \centering
  \includegraphics[width=10cm,height=5cm]{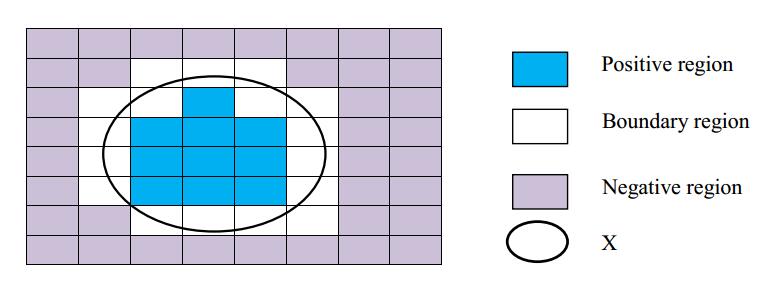}\\
  \caption{Rough set diagrammatic sketch}
\end{figure}

$\underline{R_{B}}X$ means the basic concepts of $U/R_{B}$ can be certainly assigned to \emph{X}; $\overline{R_{B}}X$ means the basic concepts of $U/R_{B}$ can be possibly assigned to \emph{X}; $BN_{B}(X)$ means we can not sure whether the basic concepts of \emph{U} belong to \emph{X} or not. If the boundary region of \emph{X} is empty which indicates $BN_{B}(X)=\varnothing$, so the set \emph{X} is crisp; if $BN_{B}(X)\neq \varnothing$, the set \emph{X} is rough \cite{CHEN2017226}. When the attributes contain more knowledge, the crisper the concepts will be; boundary region represents the uncertainty degree of knowledge; the greater the boundary is, the greater the uncertainty of knowledge will be \cite{MA201540}. In order to measure the uncertainty of rough set, the following definition is introduced.

\begin{myDef}
\normalfont{\cite{ZHANG2016323} For an information system $IS=\left \langle U,A,V,f \right \rangle  $, $A=C\cup D$, and \emph{C} is the condition attributes set, \emph{D} is the decision attribute set. $\forall B\subseteq C$, so the approximate quality of \emph{B} for \emph{D} is defined as:
\begin{equation}
  \gamma _{B}(D) =\frac{\left | POS_{B}(D) \right |}{\left | U \right |}=\frac{\left | \cup _{X\in U/D}\underline{R_{B}}X \right |}{\left | U \right |}
\end{equation}
Approximate precision is as:
\begin{equation}
  \alpha_{B}(D)=\frac{\left | POS_{B}(D) \right |}{\sum\limits_{X\in U/D}\left | \overline{B}(X) \right |}
\end{equation}
Where $U/D=\left \{ X_{1}, X_{2},\cdots, X_{n}\right \}, X_{i}\cap X_{j}=\varnothing (\forall i\neq j), U=\cup _{i=1}^{n}X_{i}$.
}
\end{myDef}
$\gamma _{B}(D)$ is also called as dependence degree of \emph{B} on \emph{D}. If $\gamma _{B}(D)=1$, it says \emph{B} is completely dependent on \emph{D}. If $0<\gamma _{B}(D)< 1$, \emph{B} is partial dependent on \emph{D}.

\begin{myDef}
\normalfont{\cite{YAO2017601} Give a decision table $DT=\left \langle U,C,D \right \rangle $, \emph{C} is the condition attributes set, \emph{D} is the decision attribute set, and $B\subseteq C$. For $\forall a\in B $, if $POS_{B}(D)=POS_{B-\left \{ a \right \}}$, it is said that \emph{a} is redundant for \emph{D}; otherwise a is essential for \emph{D}.
}
\end{myDef}

It is obvious that removing the redundant attributes relative to \emph{D} in \emph{B} will not change the approximate ability of \emph{B} to \emph{D}.
\section{The model of RELM}
In this section, we give an introduction about the structure of RELM, and then describes the basic principles of RELM. How to train hidden nodes of RELM using rough set theory can be also found in this section. The detailed execution steps of RELM can be seen in \textbf{Algorithm 3}.
\subsection{The structure of RELM}
RELM is a development of ELM \cite{Huang2006Extreme}, but different from the conventional ELM, the neurons of RELM are rough neurons which are trained by the data divided by rough set. Rough set divide a universe into two distinct parts: lower approximation set and upper approximation set. For RELM, each neuron contains two neurons: upper approximation neuron which is trained by upper approximation set and lower approximation neuron which is trained by lower approximation set. The input weights and biases of upper approximation neurons and lower approximation neurons are randomly generated. The output weights of upper approximation neurons and lower approximation neurons are analytically determined as the method of ELM. The classification result of RELM is decided by the outputs of upper approximation neurons and lower approximation neurons although the training process of these two kinds of neurons is relatively independent. It is known that RELM closely combines ELM with rough sets and the division result of a universe by rough set is used to guide the learning process of RELM. So RELM is a kind of ELM based on uncertainty measure and can effectively analyze imprecise, inconsistent and incomplete information. The structure of RELM is showed as Fig.2.
\begin{figure}
  \centering
  \includegraphics[width=12cm,height=10cm]{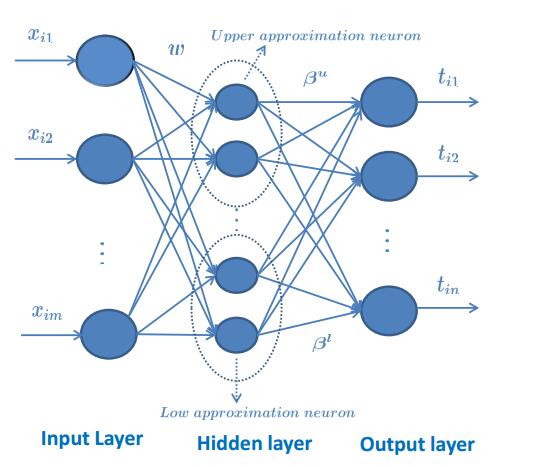}\\
  \caption{The structure of RELM}
\end{figure}

\subsection{Rough extreme learning machine for data classification}
In the algorithm of RELM, the most important is training rough neurons. From Fig.2, it is known that there are two kinds of neurons: lower approximation neuron and upper approximation neuron. Each neuron actually contains one lower approximation neuron and one upper approximation neuron. For a training data set $\mathcal{X}$, \emph{A} is the condition attributes set and \emph{D} is the decision attribute set. So $\mathcal{X}$ is divided into two parts by rough set: lower approximation set $\mathcal{X}_{lower}$ and upper approximation set $\mathcal{X}_{upper}$.
\begin{equation}
  \mathcal{X}_{lower}=POS_{B}(D)=\cup_{Y\in \mathcal{X}/D}\underline{R_{B}}Y\ \ \ \ \text{and} \ \ \ \ \mathcal{X}_{upper}=\overline{R_{B}}\mathcal{X}
\end{equation}
Let \emph{L} is the number of neurons in hidden layer, $\bm{w}_{l}$ is the input weights connecting input neurons with lower approximation neurons and $b_{l}$ is the biases of lower approximation neurons. Lower approximation neurons of RELM are trained by $\mathcal{X}_{lower}$. According to the ELM theory, if $\emph{L}\leq N_{l}$, the output weights of lower approximation neurons are as:
\begin{equation}
  \bm{\beta}^{l}=\bm{H}_{l}^{\dagger }\bm{T}_{l}=\bm{H}_{l}^{T}({\frac{\bm{I}}{C}+\bm{H}_{l}\bm{H}_{l}^{T}})^{-1}\bm{T}_{l}
\end{equation}
Where $N_{l}$ is the number of samples in $\mathcal{X}_{lower}$, \emph{C} is a ridge parameter, $\bm{H}_{l}$ is the output matrix of lower approximation neurons in hidden layer, $\bm{T}_{l}$ is the target matrix of $\mathcal{X}_{lower}$ and $\bm{I}$ is an identity matrix. $\bm{H}_{l}$ and $\bm{T}_{l}$ in Eq.(22) are as follows:

\begin{equation}
  \bm{H}_{l}=
  \begin{bmatrix}
 g(\bm{w}_{1}^{l},b_{1}^{l},\bm{x}_{1}^{l})&\cdots   & g(\bm{w}_{L}^{l},b_{L}^{l},\bm{x}_{1}^{l})\\
 \vdots & \ddots  & \vdots \\
 g(\bm{w}_{1}^{l},b_{1}^{l},\bm{x}_{N_{l}}^{l})&\cdots   & g(\bm{w}_{L}^{l},b_{L}^{l},\bm{x}_{N_{l}}^{l})
\end{bmatrix}_{N_{l}\times L}
\ \ \text{and}\ \
\bm{T}_{l}=\begin{bmatrix}
\bm{t}_{1}^{T}\\
\vdots \\
\bm{t}_{N_{l}}^{T}
\end{bmatrix}_{N_{l}\times n}
\end{equation}
If $\emph{L}> N_{l}$, the output weights of lower approximation neurons are as:
\begin{equation}
  \bm{\beta}^{l}=H_{l}^{\dagger }\bm{T}_{l}=({\frac{\bm{I}}{C}+\bm{H}_{l}^{T}\bm{H}_{l}})^{-1}\bm{H}_{l}^{T}\bm{T}_{l}
\end{equation}
Let $\bm{w}_{u}$ is the input weights connecting input neurons with upper approximation neurons and $b_{u}$ is the biases of upper approximation neurons. Upper approximation neurons of RELM are trained by $\mathcal{X}_{upper}$. So if $\emph{L}\leq N_{u}$, the output weights of upper approximation neurons are as:
\begin{equation}
  \bm{\beta}^{u}=\bm{H}_{u}^{\dagger }\bm{T}_{u}=\bm{H}_{u}^{T}({\frac{\bm{I}}{C}+\bm{H}_{u}\bm{H}_{u}^{T}})^{-1}\bm{T}_{u}
\end{equation}
Where $N_{u}$ is the number of samples in upper approximation set. $\bm{H}_{u}$ and $\bm{T}_{u}$ are as:
\begin{equation}
  \bm{H}_{u}=
  \begin{bmatrix}
 g(\bm{w}_{1}^{u},b_{u}^{l},\bm{x}_{1}^{u})&\cdots   & g(\bm{w}_{L}^{u},b_{L}^{u},\bm{x}_{1}^{u})\\
 \vdots & \ddots  & \vdots \\
 g(\bm{w}_{1}^{u},b_{1}^{u},\bm{x}_{N_{u}}^{u})&\cdots   & g(\bm{w}_{L}^{u},b_{L}^{u},\bm{x}_{N_{u}}^{u})
\end{bmatrix}_{N_{u}\times L}
\ \ \text{and}\ \
\bm{T}_{u}=\begin{bmatrix}
\bm{t}_{1}^{T}\\
\vdots \\
\bm{t}_{N_{u}}^{T}
\end{bmatrix}_{N_{u}\times n}
\end{equation}
If $\emph{L}> N_{u}$, the output weights of lower approximation neurons are as:
\begin{equation}
  \bm{\beta}^{u}=H_{u}^{\dagger }\bm{T}_{u}=({\frac{\bm{I}}{C}+\bm{H}_{u}^{T}\bm{H}_{u}})^{-1}\bm{H}_{u}^{T}\bm{T}_{u}
\end{equation}
Suppose the test data set is $\mathcal{Y}_{test}$, $\bm{h}_{lower}$ and $\bm{h}_{upper}$ are the output matrices of lower approximation neurons and upper approximation neurons for $\mathcal{Y}_{test}$ correspondingly. The final output matrices of hidden layer are as:
\begin{equation}
  \bm{h}_{upper} = \max(\bm{h}_{lower},\bm{h}_{upper}) \ \ \ \text{and} \ \ \
  \bm{h}_{lower} = \min(\bm{h}_{lower},\bm{h}_{upper})
\end{equation}
So the target matrix of RELM for $\mathcal{Y}_{test}$ is
\begin{equation}
  \bm{T}_{test}=c\cdot\bm{T}_{test}^{lower}+(1-c)\cdot\bm{T}_{test}^{upper}=c\cdot\bm{h}_{lower}\bm{\beta}^{l}+(1-c)\cdot\bm{h}_{upper}\bm{\beta}^{u}
\end{equation}
Where \emph{c} is a weight to balance the output matrices of lower approximation neurons and upper approximation neurons. $\bm{T}_{test}^{lower}$ is the output target matrix of lower approximation neurons and $\bm{T}_{test}^{upper}$ is the output target matrix of upper approximation neurons.

In order to eliminate the influence of redundant attributes on classification result, attribute reduction is introduced in RELM. From rough set theory, it is known that rough set can remove redundant attributes without any empirical knowledge. For the training data set $\mathcal{X}$, \emph{C} is the condition attributes set, \emph{D} is the decision attribute set and $B\subseteq C$; $\forall a\in C-B$, the significance of the attribute \emph{a} for the decision attribute set \emph{D} based on the condition attributes set \emph{B} is as:
\begin{equation}
  sig(a,B,D)=\gamma _{B \cup\left \{ a \right \}}(D)-\gamma _{B}(D)
\end{equation}
The greater the value of $sig(a,B,D)$ is, the more important \emph{a} is for \emph{D}. So $sig(a,B,D)$ can be used to select non-redundant attributes. The attribute reduction method is as follow.
\begin{algorithm} \caption{ Attributes reduction.}
  \begin{algorithmic}[1]
   \Require
    A decision table $DT=\left \langle \mathcal{X}, C, D \right \rangle$
    \Ensure Relative reduction $B$.
    \State $B=\varnothing$;
    \State Calculate the approximate quality of \emph{C} for $\emph{D}: \gamma _{C}(D)$;
    \For{each $a\in C-B \ \& \ C-B \neq \varnothing$}
    \State Calculate $sig(a,B,D)$;
    \EndFor
    \State Select an attribute $a$ with maximum $sig(a,B,D)$;
    \If{$\gamma _{C- \left \{ a \right \}}(D)-\gamma _{C}(D)<0$}
    \State $B=B\cup \left \{ a \right \}$;
    \State Go to $\bm{step 3}$;
    \EndIf
    \For{$a\in B \ \& \ B\neq \varnothing$}
    \If{$\gamma _{B-\left \{ a \right \} }(D)-\gamma _{B}(D)\geq0$}
    \State $B=B-\left \{ a \right \}$;
    \EndIf
    \EndFor
    \State Output the attributes set \emph{B};
   \end{algorithmic}
\end{algorithm}

The number of neurons in the hidden layer plays an important role in RLM. If the number of neurons in the hidden layer \emph{L} of RELM is too large, it may be overfitting; if \emph{L} is too small, the under fitting problem may appear. So RELM uses the dividing result of data by rough set to determine the number of neurons in the hidden layer. For a data set, if the larger the positive region is, it presents that the attributes set has a good ability to distinguish data, so it is better for RELM to determine a small \emph{L}; if the larger the boundary region is, it indicates the attributes set can not divide data well, and it has a trend to choose a large \emph{L}. The number of neurons in the hidden layer \emph{L} is decided as
\begin{equation}
  L=k_{1}\cdot\frac{\left | \mathcal{X} \right |}{\left | POS_{B}(D) \right |} +k_{2}\cdot \frac{\left | \mathcal{X} \right |-\left |POS_{B}(D)  \right |}{\left | \mathcal{X} \right |}
\end{equation}
Where $k_{1}$ and $k_{2}$ are the parameters predefined by user which mean the weights of positive region and boundary region correspondingly for determining \emph{L}. It is obvious that \emph{L} is decided according to the division of data self, in other words, it decreases the dependence on empirical knowledge; so the number of neurons in the hidden layer is determined by Eq.(31) can reduce the blindness of selecting \emph{L} to a certain extent.

From the above descriptions, the steps of RELM are summarized as \textbf{Algorithm 3}.
\begin{algorithm} \caption{ RELM algorithm.}
  \begin{algorithmic}[1]
   \Require
    A training data set $\mathcal{X}$;
    a testing data set $\mathcal{Y}$;
    the parameters $k_{1}$, $k_{2}$ and \emph{c};
    the activation function $g(\cdot)$.
    \Ensure The classification result of $\mathcal{Y}$.
    \State \textbf{Training}:
    \State Divide the data set $\mathcal{X}$ into the two parts by rough set: upper approximation set $\mathcal{X}_{upper}$ and lower approximation set $\mathcal{X}_{lower}$;
    \State Get the reduction set \emph{red} from $\bm{step 2}$ and remove the redundant attributes from the data set: $\mathcal{X}_{upper}$, $\mathcal{X}_{lower}$ and $\mathcal{Y}$;
    \State Randomly generate the input weights and bias in hidden layer for upper approximation neurons and lower approximation neurons;
    \State Determining the number of neurons in the hidden layer \emph{L} according Eq.(31);
    \State Train upper approximation neurons using $\mathcal{X}_{upper}$ and obtain the output weights $\bm{\beta}^{u}$;
    \State Train lower approximation neurons using $\mathcal{X}_{lower}$ and obtain the output weights $\bm{\beta}^{l}$;
    \State \textbf{Testing}:
    \State Calculate the final output matrices of hidden layer of the testing data set $\mathcal{Y}$: $\bm{h}_{upper}$ and $\bm{h}_{lower}$ according to Eq.(28);
    \State Get the target output matrix of upper approximation neurons and lower approximation neurons: $\bm{T}_{test}^{upper}$ and $\bm{T}_{test}^{lower}$;
    \State Calculate the target output matrix $\bm{T}_{test}$ according to Eq.(29);
    \end{algorithmic}
\end{algorithm}

From the algorithm 3, the steps of RELM and ELM are very different from each other. The neurons of RELM are rough neurons, which are trained based on the division of data by rough set, so RELM is a classification method based on uncertainty measure; the method of uncertainty measure has a good advantage in dealing with inconsistent and incomplete information and the number of neurons in hidden layer is also decided by the information provided by rough set which does not need too much experience knowledge. By utilizing the output results of the upper approximation neurons and lower approximation neurons, RELM can make full use of the information provided by rough sets to get a better classification result.

\section{Experiment and results}
In this section, we demonstrate the effectiveness of the proposed method and comparison algorithms on 19 data sets. To verify the capabilities of our algorithm, we choose CELM, CSELM, DELM, MELM, RandomSampleELM and SELM etc as comparison algorithms, all algorithms are executed on MATLAB R2017a platform. The configurations of the computer are as: Windows 7 OS, 8GB RAM memory, Intel i3-2120 dual core CPU.
\subsection{Data set descriptions}
In the experiments, there are 16 real data sets from the UCI data sets website\footnote{http://archive.ics.uci.edu/ml/datasets.html} and 3 man-made data sets which are generated by MOA platform \footnote{http://moa.cms.waikato.ac.nz/} \cite{bifet2010moa:}. The descriptions of the 16 real data sets can be seen from the UCI data sets website, so we only give a brief introduction about the 3 man-made data sets. The information of all data sets can be seen in Table 1.

\emph{Hyperplane} data set: For a \emph{d}-dimensional space, a hyperplane is defined as $\sum\limits_{i=1}^{d} w_{i}x_{i}=w_{0}$ where $w_{0}=\frac{1}{2}\sum\limits_{i=1}^{d}w_{i}$ and $x_{i}\in \left [ -10,10 \right ]$. If $\sum\limits_{i=1}^{d} w_{i}x_{i}\geq w_{0}$, the label of $\bm{x}$ is remarked as a positive sample; otherwise the label of $\bm{x}$ is marked as a negative sample. There are 10\% noise in the \emph{Hyperplane} data set.

\emph{Waveform} data set: There are 3 classes, 21 attributes in the data set. The goal of the task is to differentiate the three types of waveform. There are 2\% noise in the \emph{Hyperplane} data set.

\emph{STAGGER} data set: There are 3 attributes for each sample in the data set: $color \in \left \{ green, blue, red \right \}$, $shape \in \left \{triangle, circle, retangle\right \}$, and $size \in \left \{ small, medium, large \right \}$. The concepts of the data are as: $color =red \wedge size=small$, $color =green \vee shape=large$ and $size=mediun \vee size=large$.

\begin{table}[h]
  \centering
  \small
  \caption{The information of the experimental data sets}
  \begin{tabular}{ccccccccccc}\toprule
  Data set &Attributes &Numerical attributes &Categorical attributes &Samples &Type\\\midrule
  Horse &26 &9 &17 &300 &mixed\\
  glass &9 &9 &0 &214 &numerical\\
  biodeg&41 &17 &24 &500 &mixed\\
  haberman &3  &0 &3 &301 &numerical\\
  lungcancer &31 &31 &0 &57 &categorical\\
  votes &6 &6 &0 &435 &categorical\\
  Germany &24 &3 &21 &500 &mixed\\
  Echocardiogra &12 &4 &8 &131 &mixed\\
  Tic &9 &0 &9 &958 &categorical\\
  parkinsons &22 &22 &0 &195 &numerical\\
  yeast &8 &8 &0 &500 &numerical\\
  vehicle &18 &18 &0&846 &numerical\\
  pima &8 &1 &7 &500 &mixed\\
  segment&19 &19 &0 &500 &numerical\\
  Hepatitis&19 &1 &18 &156 &mixed\\
  STAGGER &3 &0 &3 &500 &categorical\\
  adult &13 &5 &8 &500 &mixed\\
  Hyperplane &40 &40 &0 &500 &numerical\\
  Wavefrom &21 &21 &0 &500 &numerical\\
  \bottomrule
\end{tabular}
\end{table}

Because rough set can only analyse categorical data, the numerical data sets and fixed data sets are discretized. The discretization method is equal interval discretization, and the number of intervals is set as the number of labels in data set.

\subsection{The comparison results of RELM with other ELM algorithms}
In order to test the efficiency of RELM, in this section, we choose CELM\cite{Zhu2014Constrained}, CSELM, DELM, MELM, RandomSampleELM and SELM as comparison algorithms\cite{Zhuwentao}. For RELM, \emph{c}=0.5; the activation function is chosen from \emph{sigmoid, radbas, tribas, sine} and \emph{hardlim}. The other parameters of RELM and comparison algorithms are as in Tables 2 and 3. The test results and time overheard are also showed in Tables 2 and 3.
\begin{sidewaystable}
  \centering
  \caption{The test results of RELM and comparison algorithms on the data sets}
  \scriptsize
  \begin{tabular}{ccccccccccc}\toprule
        & CELM & CSELM & DELM & MELM & RandSampleELM & SELM & RELM & L & C & function \\\midrule
  Horse & 0.6470$\pm$0.02163 & 0.0970$\pm$0.0221 &0.6650$\pm$0.0877 & \textbf{0.6680$\pm$0.1223} & 0.4630$\pm$0.1671 & 0.6480$\pm$ 0.00632 &0.6550$\pm$0.0513 & 2 & 100 & hardlim \\
  glass & 0.7430$\pm$0.0219 & 0.0980$\pm$0.0225 & \textbf{0.7530$\pm$0.1011} &0.7450$\pm$0.1032 & 0.7410$\pm$0.3956 & 0.6450$\pm$0.0401 &0.6160$\pm$0.0474& 20 & 100 &hardlim\\
  biodeg & 0.6537$\pm$0.0163 & 0.0437$\pm$0.0033 & 0.6559$\pm$0.0127 & 0.6605$\pm$0.0245 & 0.4846$\pm$0.5076 & \textbf{0.6690$\pm$0.0057} & 0.6614$\pm$0.0913 & 5 & 100 & hardlim \\
  haberman & 0.7188$\pm$0.0069 & 0.0317$\pm$0.0160 & 0.7128$\pm$0.0104 & 0.7128$\pm$0.0224 & 0.6227$\pm$0.4778 & 0.6940$\pm$0.0031 & \textbf{0.7435$\pm$0.0490} & 5 & 100 & tribas \\
  lungcancer & 0.3818$\pm$0.0575 & \textbf{1$\pm$0} & 0.4364$\pm$0.0717 & 0.3636$\pm$0.1212 & 0.4363$\pm$0.1858 & 0.3636$\pm$0.0383 & 0.3182$\pm$0.1154 & 20 & 100 &tribas \\
  votes & 0.7745$\pm$0.0298 & 0.6579$\pm$0.0022 & 0.7965$\pm$0.0327 & 0.9276$\pm$0.01810 & 0.5489$\pm$0.0146 & 0.9331$\pm$0.0033 & \textbf{0.9355$\pm$0.1044} & 1000 & 100 & radbas \\
  Germany & 0.6874$\pm$0.0025 & 0.4131$\pm$0.0101 & 0.6841$\pm$0.0047 & 0.6832$\pm$0.0019 & 0.3892$\pm$0.1956 & 0.6805$\pm$0.0104 & \textbf{0.6958$\pm$0.0197} & 2 & 1000 & tribas \\
  Echocardiogram & 0.7386$\pm$0.0482 & 0.0113$\pm$0.0120 & 0.7227$\pm$0.0869 & 0.6795$\pm$0.0783 & 0.1977$\pm$0.3127 & 0.6727$\pm$0.0618 & \textbf{0.7409$\pm$0.0565} & 2 & 1000 & tribas \\
  Tic & 0.6571$\pm$0.0179 & 0.4934$\pm$0.0252 & 0.6475$\pm$0.0243 & 0.6553$\pm$0.0190 & 0.4356$\pm$0.0190 & 0.6559$\pm$0.0154 & \textbf{0.6603$\pm$0.0114} & 5 & 1000 &sigmoid \\
  parkinsons & 0.8061$\pm$0.050 & 0$\pm$0 & 0.8076$\pm$0.0424 & 0.7769$\pm$0.0621 & 0.1692$\pm$0.3458 & 0.7676$\pm$0.0419 & \textbf{0.8169$\pm$0.0816} & 800 & 1000 & hardlim \\
  yeast & 0.5400$\pm$0.0189 & 0.0255$\pm$0.0079 & 0.5478$\pm$0.0307 & 0.6791$\pm$0.0156 & 0.5640$\pm$0.1990 & \textbf{0.7058$\pm$0.0209} & 0.6024$\pm$0.1589 & 1000 & 1000 & hardlim \\
  vehicle & 0.7322$\pm$0.1120 & 0.0716$\pm$0.0119 & 0.7478$\pm$0.0674 & 0.7457$\pm$0.0641 & 0.5042$\pm$0.4190 & 0.7592$\pm$0.0188 & \textbf{0.7691$\pm$0.0217} & 2 & 1000 &radbas \\
  pima &0.6609$\pm$0.0320 & 0.1625$\pm$0.0235 & 0.6742$\pm$0.0461 & 0.6656$\pm$0.0409 & 0.2613$\pm$0.1118 & 0.6367$\pm$0.0276 & \textbf{0.6820$\pm$0.0201} & 5 & 1000 & tribas \\
  segment & 0.8551$\pm$0.0209 & 0.0239$\pm$0.0085 & 0.8551$\pm$0.0304 & 0.8616$\pm$0.0282 & 0.3233$\pm$0.3480 & 0.8598$\pm$0.0303 & \textbf{0.8658$\pm$0.0302} & 5 & 1000 & hardlim \\
  Hepatitis & 0.6807$\pm$0.0553 & 0.1134$\pm$0.0400 & 0.6865$\pm$0.0588 & 0.7692$\pm$0.0351 & 0.6135$\pm$0.1987 & 0.6653$\pm$0.0943 & \textbf{0.7769$\pm$0.0553} & 100 & 1000 &tribas \\
  STAGGER&0.9898$\pm$0.0057 &0.6203$\pm$0.0328 &\textbf{1$\pm$0} &0.5065$\pm$0.0360 &0.6281$\pm$0.0351 &0.5059$\pm$0.0251 &\textbf{1$\pm$0} &100 &1000 &hardlim\\
  adult&0.7197$\pm$0.0297 &0.0161$\pm$0.0080 &0.6880$\pm$0.0396 &0.7293$\pm$0.0203 &0.6024$\pm$0.5134 &0.7491$\pm$0.0356 &\textbf{0.7682$\pm$0.0217} &100 &1000 &tribas\\\bottomrule
\end{tabular}
\end{sidewaystable}

\begin{table}[h]
  \centering
  \caption{The time overheard of RELM and comparison algorithms on the data sets}
  \footnotesize
  \begin{tabular}{ccccccccccc}\toprule
        & CELM & CSELM & DELM & MELM & RandSampleELM & SELM & RELM \\\midrule
  Horse & 0.0147 & 0.0087 & 0.0091 & 0.0099 & 0.0156 & 0.0087 & 18.0397 \\
  glass & 0.0033 & 0.0629 & 0.0051 & 0.0092 & 0.0046 & 0.0080 & 0.2921 \\
  biodeg& 0.0072 & 0.3173 & 0.0119 & 0.0142 & 0.0188 & 0.0059 & 90.9761 \\
  haberman &0.0072  & 0.0191 & 0.0053 & 0.0059 &0.0073 & 0.0043 & 0.035092\\
  lungcancer & 0.0031 & 0.0027 & 0.0033 & 0.0036 & 0.0048 & 0.0033 & 10.5244 \\
  votes & 0.8055 & 0.1512 & 0.1584 & 0.1735 & 2.9999 & 0.1204 & 4.0429 \\
  Germany & 0.0123 & 0.0091 & 0.0083 & 0.0164 & 0.0095 & 0.0090 & 39.0233 \\
  Echocardiogram & 0.0020 & 0.0107 & 0.0033 & 0.0029 & 0.0020 & 0.0026 & 0.6670 \\
  Tic & 0.0092 & 0.0563 & 0.0087 & 0.0085 & 0.0108 & 0.0087 & 3.435405 \\
  parkinsons & 0.0792 & 0.0204 & 0.0621 & 0.0916 & 1.9678 & 0.0508 & 2.3083 \\
  yeast & 0.2022 & 0.3253 & 0.2162 & 0.1776 & 2.9894 & 0.1848 & 0.5858 \\
  vehicle & 0.0068 & 0.0142 & 0.0073 & 0.0075 & 0.0098 & 0.0088 & 3.7902 \\
  pima & 0.0078 & 0.0091 & 0.0085 & 0.0073 & 0.0100 & 0.0062 & 0.0059 \\
  segment & 0.0069 & 0.2842 & 0.0069 & 0.0050 & 0.0082 & 0.0072 & 3.0836 \\
  Hepatitis &0.0089  &0.0142  &0.0076  &0.0078  &0.0407  &0.0069  &4.9032  \\
  STAGGER&0.0095 &0.0099 &0.0082 &0.0141 &0.0358 &0.0114 &0.0407\\
  adult&0.0107 &0.0432 &0.0092 &0.0103 &0.0382 &0.0087 &3.3013\\\bottomrule
\end{tabular}
\end{table}

From Table 2, it is obvious that RELM is better than CELM, CSELM, DELM, MELM, RandSampleELM, SELM and RELM on most data sets; RELM gets the highest accuracy on 12 data sets and gets the second best on \emph{biodeg} data set; it only loses to DELM, CSELM, MELM and SELM on \emph{Horse}, \emph{glass}, \emph{biodeg}, \emph{lungcncer}, and \emph{yeast} data sets; the results indicate that the approximation ability of RELM is effective for classification task. Table 3 is the time overheard of RELM and the comparison algorithms. By analyzing the data, it is known that the proposed algorithm does not have much advantage on time overheard and it is not the least time-consuming algorithm on most data sets, in other words, RELM is a time-consuming algorithm. According to the ELM theory, ELM has a fast speed, so the most time is consumed in the rough set method. If combining the data from Table 2, it shows that rough set can improve the performance of the proposed algorithm; therefore for the classification task with low real-time requirement, it is worth considering using RELM.

\subsection{The effect of activation function on the performance of the algorithm}
For testing the effect of activation function on the performance of RELM, we choose \emph{sigmoid}, \emph{radbas}, \emph{tribas}, \emph{sine}, and \emph{hardlim} as activation functions. Every data set is tested on the 5 activation functions and RELM is tested on 17 data sets. The number of neurons and test results are showed in Table 4.
\begin{table}[h]
  \centering
  \caption{The test results of RELM with different activation functions}
  \footnotesize
  \begin{tabular}{ccccccccccc}\toprule
        & sigmoid & radbas & tribas & sine & hardlim & L \\\midrule
  Horse &\textbf{0.6650$\pm$0.0375}  &0.5900$\pm$0.0956  &0.6510$\pm$0.0387  &0.6210$\pm$0.0927  &0.6580$\pm$0.0537  & 20 \\
  glass &0.8808$\pm$0.0329  &0.8644$\pm$0.0332  &0.8575$\pm$0.0377  &0.6986$\pm$0.3069  &\textbf{0.8836$\pm$0.0378}  & 90\\
  biodeg&\textbf{0.5934$\pm$0.0351}  &0.5341$\pm$0.0669  &0.5599$\pm$0.0895  &0.5719$\pm$0.1030  &0.5664$\pm$0.07752  &100 \\
  haberman &0.6683$\pm$0.1556  &0.6228$\pm$0.1969  &\textbf{0.7297$\pm$0.0371}  &0.5485$\pm$0.1089  &0.7267$\pm$0.0234  &50 \\
  lungcancer &\textbf{0.3636$\pm$0.1050}  &0.3091$\pm$0.1497  &0.29091$\pm$0.0939  &0.3000$\pm$0.1425  &0.2818$\pm$0.1246  &100  \\
  votes &\textbf{0.6296$\pm$0.1351}  &0.5807$\pm$0.0657  &0.5800$\pm$0.1611  &0.5552$\pm$0.0696  &0.5676$\pm$0.1328  &20  \\
  Germany &0.6985$\pm$0.0171  & 0.6153$\pm$0.1639 &0.6686$\pm$0.0988  &0.5153$\pm$0.0987  &\textbf{0.7021$\pm$0.0148}  &30 \\
  Echocardiogram &0.5909$\pm$0.1680  &0.6500$\pm$0.1189  &0.6909$\pm$0.0559  &0.4796$\pm$0.1156  &\textbf{0.7205$\pm$0.0372}  &50 \\
  Tic &0.6591$\pm$0.0289  &0.6525$\pm$0.0143  &\textbf{0.6653$\pm$0.0161}  &0.5844$\pm$0.1329  &0.6528$\pm$0.0115  &50  \\
  parkinsons &\textbf{0.7431$\pm$0.0513}  &0.3862$\pm$0.2113  &0.2231$\pm$0.0364  &0.5477$\pm$0.1395  &0.6046$\pm0.2803$  &30  \\
  yeast &0.7413$\pm$0.0266  &0.7365$\pm$0.0385  &0.7270$\pm$0.0405  &0.6755$\pm$0.1635  &\textbf{0.7497$\pm$0.0211}  &100 \\
  vehicle &0.7461$\pm$0.0203  &0.7379$\pm$0.0260  &\textbf{0.7507$\pm$0.0197}  &0.5766$\pm$0.1410  &0.7415$\pm$0.0231  &25  \\
  pima &0.5850$\pm$0.1197  &0.6204$\pm$0.0450  &\textbf{0.6437$\pm$0.0279}  &0.4970$\pm$0.0881  &0.6383$\pm$0.0323  &75 \\
  segment &0.8689$\pm$0.0192  &0.8593$\pm$0.0273  &\textbf{0.8695$\pm$0.0292}  &0.4479$\pm$0.2288  &0.8665$\pm$0.0312  &45 \\
  Hepatitis &\textbf{0.8192$\pm$0.0603}  &0.8154$\pm$0.0427  &0.8135$\pm$0.0301  &0.7058$\pm$0.2277  &0.4213$\pm$0.2840  &80 \\
  STAGGER &0.5539$\pm$0.0971  &0.4808$\pm$0.02526  &0.5222$\pm$0.1179  &\textbf{0.6647$\pm$0.1967}  &0.4773$\pm$0.0536  &50 \\
  adult   &\textbf{0.7767$\pm$0.0245}  &0.7713$\pm$0.0359  &0.7689$\pm$0.0287  &0.5359$\pm$0.1859  &0.7731$\pm$0.0252  &50 \\\bottomrule
\end{tabular}
\end{table}

From Table 4, it can conclude that the accuracies of RELM are different with different activation functions. When the activation function is \emph{sigmoid}, RELM gets 7 best accuracies; when the activation function is \emph{tribas}, RELM gets 5 best accuracies; when the activation function is \emph{hardlim}, RELM gets 4 best accuracies; the test results of the proposed algorithm are not very well for those experimental data sets if choosing \emph{radbas} or \emph{sine} as activation function, because there is no best accuracy for \emph{radbas} and there is only one best accuracy for \emph{sine}. For every data set, the standard deviations of accuracies are also different with activation function. For example, the activation function is \emph{sigmoid}, the standard deviation is 0.0375; if the activation function is \emph{radbas}, it changes to 0.0956. the similar situations can be found on the other data sets. From the test results, it is obvious that the activation function has a great impact on the performance of RELM; the activation functions of the best results on different data sets are also different, so how to select activation function depending on experimental data set. If users do not have too much empirical knowledge about choosing activation functions, \emph{sigmoid}, \emph{tribas} or \emph{hardlim} seems be a good initial selection for those experimental data sets.

\subsection{The effect of the number of neurons in hidden layer on the performance of RELM}
In order to test the effect of the number of nodes in hidden layer on performance of RELM, we set the activation function of RELM as \emph{sigmoid}, the ridge parameter \emph{C} is set as 1000 and the number of neurons in hidden layer  varies from 1 to 1000. The test accuracies of RELM with different number of the neurons in hidden layer are showed as in Table 5.

\begin{table}[h]
  \centering
  \caption{The test results of RELM with different numbers of neurons in hidden layer}
  \scriptsize
  \begin{tabular}{ccccccccccc}\toprule
    L    & 5 & 50 & 200 &500 & 800 &1000\\\midrule
  Horse  &\textbf{0.6100$\pm$0.1505} &0.5370$\pm$0.1245 &0.4640$\pm$0.1342 &0.4770$\pm$0.1084 &0.5040$\pm$0.1424 &0.5030$\pm$0.1216 \\
  glass  &0.7246$\pm$0.3046 &\textbf{0.7739$\pm$0.2346} &0.5562$\pm$0.3631 &0.4836$\pm$0.3667 &0.4972$\pm$0.3663 &0.5507$\pm$0.3667 \\
  biodeg &0.5389$\pm$0.1559 &\textbf{0.5503$\pm$0.1160} &0.4904$\pm$0.1169 &0.4868$\pm$0.0945 &0.4353$\pm$0.0727 &0.5072$\pm$0.1196 \\
  haberman &0.5891$\pm$0.1929 &\textbf{0.6267$\pm$0.2079} &0.5069$\pm$0.1772 &0.5356$\pm$0.1924 &0.4970$\pm$0.2082 &0.4327$\pm$0.1956 \\
  lungcancer &0.2000$\pm$0.0717 &0.2818$\pm$0.1512 &0.3091$\pm$0.0636 &0.3909$\pm$0.0963 &0.2636$\pm$0.1572 &\textbf{0.3455$\pm$0.1032} \\
  votes &\textbf{0.6779$\pm$0.2021} &0.6207$\pm$0.2469 &0.6269$\pm$0.2145 &0.5069$\pm$0.2220 &0.5910$\pm$0.2294 &0.6227$\pm$0.2320 \\
  Germany &0.5264$\pm$0.2160 &\textbf{0.6701$\pm$0.1807} &0.5653$\pm$0.1987 &0.5599$\pm$0.1573 &0.4491$\pm$0.1947 &0.4719$\pm$0.1459 \\
  Echocardiogram &\textbf{0.7318$\pm$0.1354} &0.6182$\pm$0.2096 &0.6386$\pm$0.1447 &0.6091$\pm$0.1827 &0.5955$\pm$0.2221 &0.6114$\pm$0.2745 \\
  Tic &0.6563$\pm$0.0304 &0.6531$\pm$0.0256 &0.6500$\pm$0.0177 &\textbf{0.6588$\pm$0.0311} &0.6550$\pm$0.0287 &0.6438$\pm$0.0231 \\
  parkinsons &\textbf{0.7569$\pm$0.0422} &0.5677$\pm$0.2581 &0.4339$\pm$0.2316 &0.5785$\pm$0.2181 &0.5431$\pm$0.1860 &0.5369$\pm$0.216 \\
  yeast &0.4635$\pm$0.2474 &\textbf{0.6994$\pm$0.1549} &0.6222$\pm$0.2063 &0.6713$\pm$0.1699 &0.5922$\pm$0.2214 &0.5431$\pm$0.2452 \\
  vehicle &\textbf{0.7545$\pm$0.0238} &0.5006$\pm$0.2552 &0.6246$\pm$0.2233 &0.6515$\pm$0.2149 &0.6060$\pm$0.2230 & 0.5497$\pm$0.2287\\
  pima &\textbf{0.5904$\pm$0.0765} &0.5246$\pm$0.1099 &0.4683$\pm$0.1204 &0.5389$\pm$0.1197 &0.541$\pm$0.1155 &0.4994$\pm$0.1026 \\
  segment &\textbf{0.8120$\pm$0.1916} &0.6761$\pm$0.2353 &0.3635$\pm$0.2873 &0.4922$\pm$0.2505 &0.6365$\pm$0.2100 &0.5150$\pm$0.2019 \\
  Hepatitis &0.7942$\pm$0.0666 &0.7808$\pm$0.0446 &\textbf{0.8173$\pm$0.0408} &0.8096$\pm$0.0492 &0.7981$\pm$0.0187 &0.8000$\pm$0.0317 \\
  STAGGER &\textbf{0.7168$\pm$0.1956} &0.5425$\pm$0.1100 &0.5150$\pm$0.0314 &0.5042$\pm$0.0176 &0.5204$\pm$0.0287 &0.5168$\pm$0.0360 \\
  adult &0.7671$\pm$0.0234 &\textbf{0.7892$\pm$0.0243} &0.7521$\pm$0.0821 &0.5467$\pm$0.2901 &0.6006$\pm$0.2756 &0.6497$\pm$0.2323 \\\bottomrule
\end{tabular}
\end{table}
From Table 5, it is known that the number of neurons in hidden layer has a significant impact on the performance of RELM. If the numbers of neurons in hidden layer are different, the test results will be also different. Because the sizes of the experimental data sets are not very large, most data sets get the best results on a small \emph{L}. By analyzing the test results, it can conclude that the performance of the proposed algorithm does not increase with the increase of the number of neurons in hidden layer. If the number of neurons in hidden layer is too large or too small, the performance of the algorithm will be decreased. The reason for this phenomenon is that if \emph{L} is too large, it will cause the structure of RELM is too complex and RLEM may be over-fitting for training data; if \emph{L} is too small, it will cause the target classification model cannot be effectively approximated by RELM and under-fitting may appear. When combining Tables 2 and 5, it can be found that RELM has a trend to choose a small \emph{L} under the condition of guaranteeing its performance; it indicates that the method determining the number of neurons in hidden layer is efficient.

\subsection{The research about the reduction mechanism of RELM}
In order to test the reduction mechanism of RELM, we execute the proposed algorithm and RELM without reduction mechanism (denoted URELM) on 14 data sets; The number of neurons in hidden layer \emph{L} is 150, the activation function is \emph{hardlim} and \emph{C} is set as 1000. RELM and URELM are tested 10 times and the results are showed in Tables 6,7 and Fig.3.

From Fig.3, it is found that the performances of RELM and URELM have a fluctuation, and the main reason is that the input weights and the biases are randomly generated but the number of neurons in hidden layer \emph{L} is fixed; randomness results in the fluctuation of RELM's performance. After analysing of the curves in Fig.3, it can see RELM is better than URELM in most cases. The results in Table 6 are the average accuracies of 10 executing results. From the results in Table 6, it is obvious that the performance of RELM is significantly better than that of URELM on all experimental data sets and the results indicate that the reduction mechanism of RELM can improve the performance of RELM. The data in Table 7 is the reduction results of RELM. From Table 7, RELM can remove redundant attributes without changing the distinguish ability of condition attributes; it can conclude that the reduction mechanism is effective for RELM. However, the reduction mechanism gets an abnormal reduction result on \emph{Hepatitis} data set. For \emph{Hepatitis} data set, there are 19 attributes, but all condition attributes are removed as redundant attributes. The reason why the reduction mechanism produces this result is that the dependence of condition attributes on decision attributes is not very large. In other words, relative positive region does not decrease when removing any condition attribute; so the reduction algorithm removed all condition attributes.

\begin{figure}[h]
  \begin{minipage}{5cm}
    \centerline{\includegraphics[width=7cm,height=4cm]{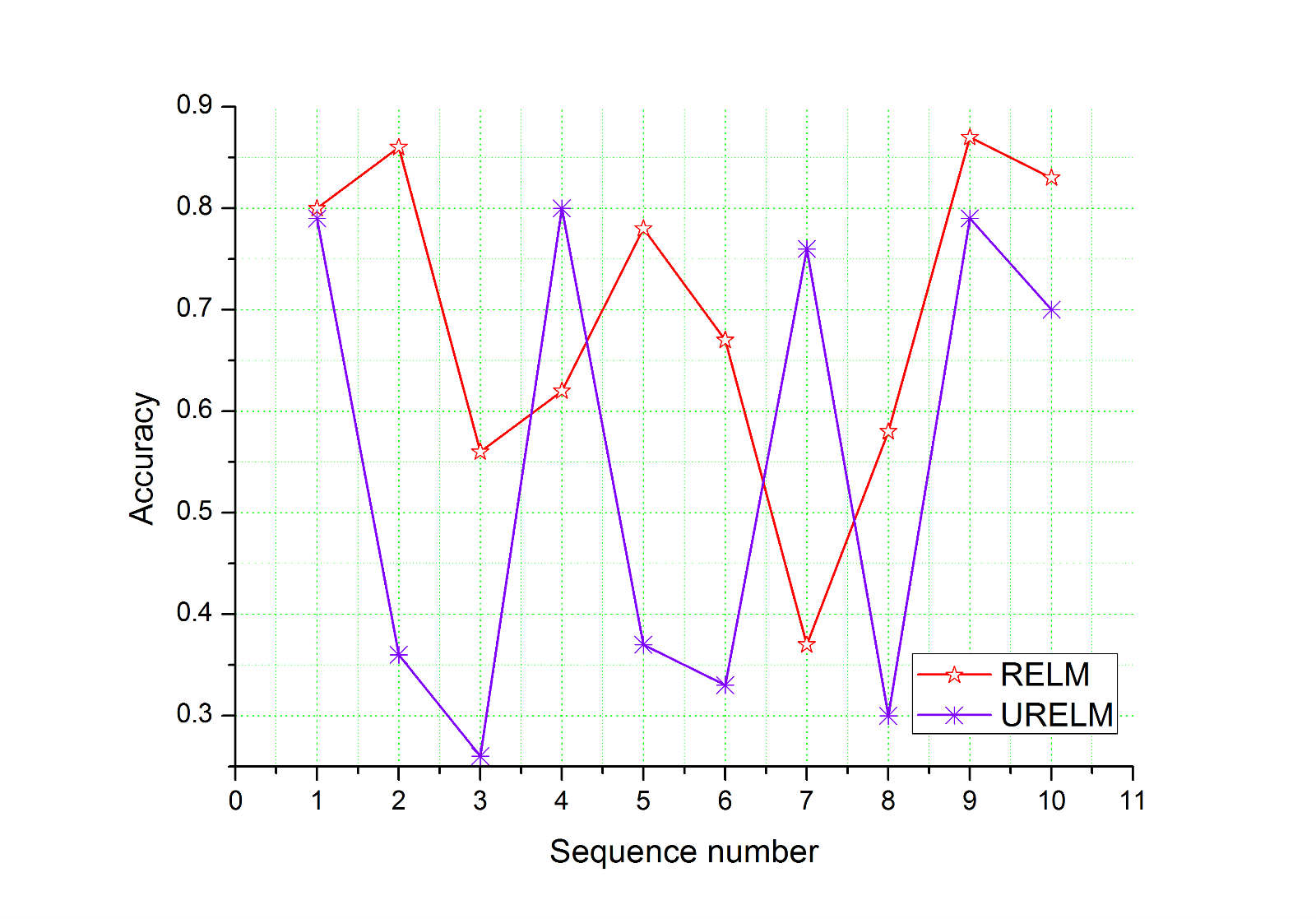}}
    \centerline{3-a Horse}
  \end{minipage}
  \hfill
  \begin{minipage}{5cm}
    \centerline{\includegraphics[width=7cm,height=4cm]{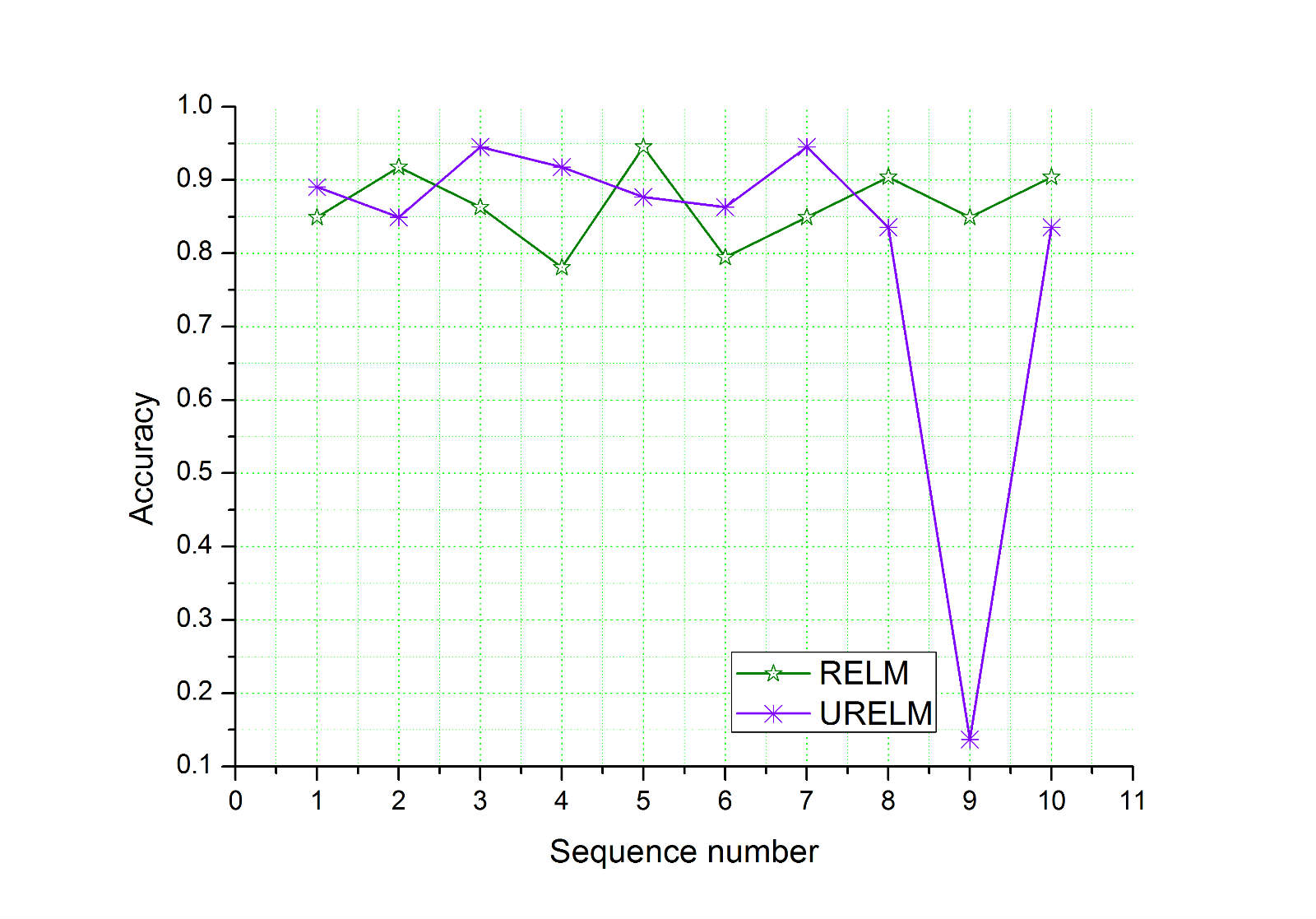}}
    \centerline{3-b glass}
  \end{minipage}
  \hfill
  \begin{minipage}{5cm}
    \centerline{\includegraphics[width=7cm,height=4cm]{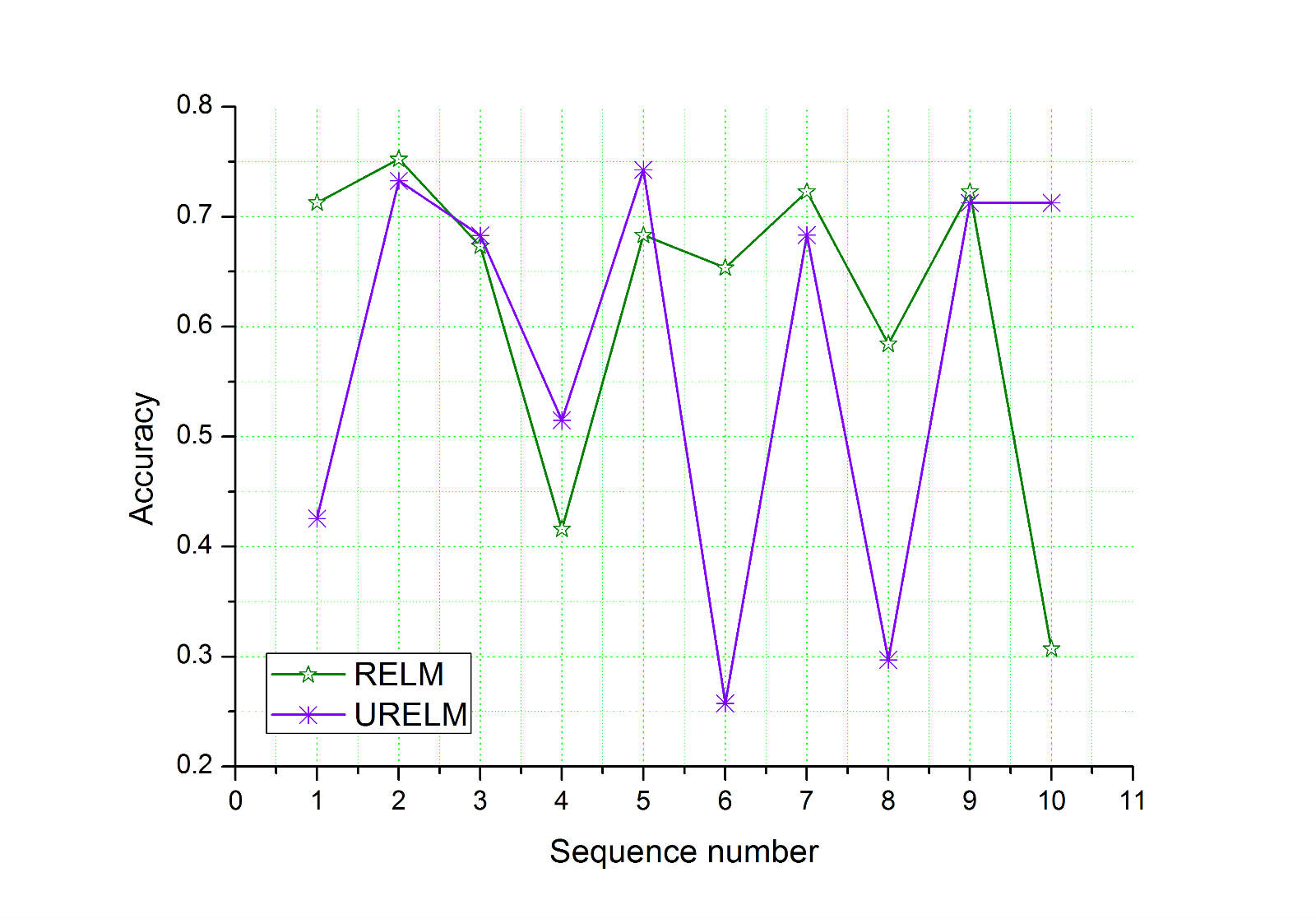}}
    \centerline{3-c haberman}
  \end{minipage}
  \vfill
  \begin{minipage}{5cm}
    \centerline{\includegraphics[width=7cm,height=4cm]{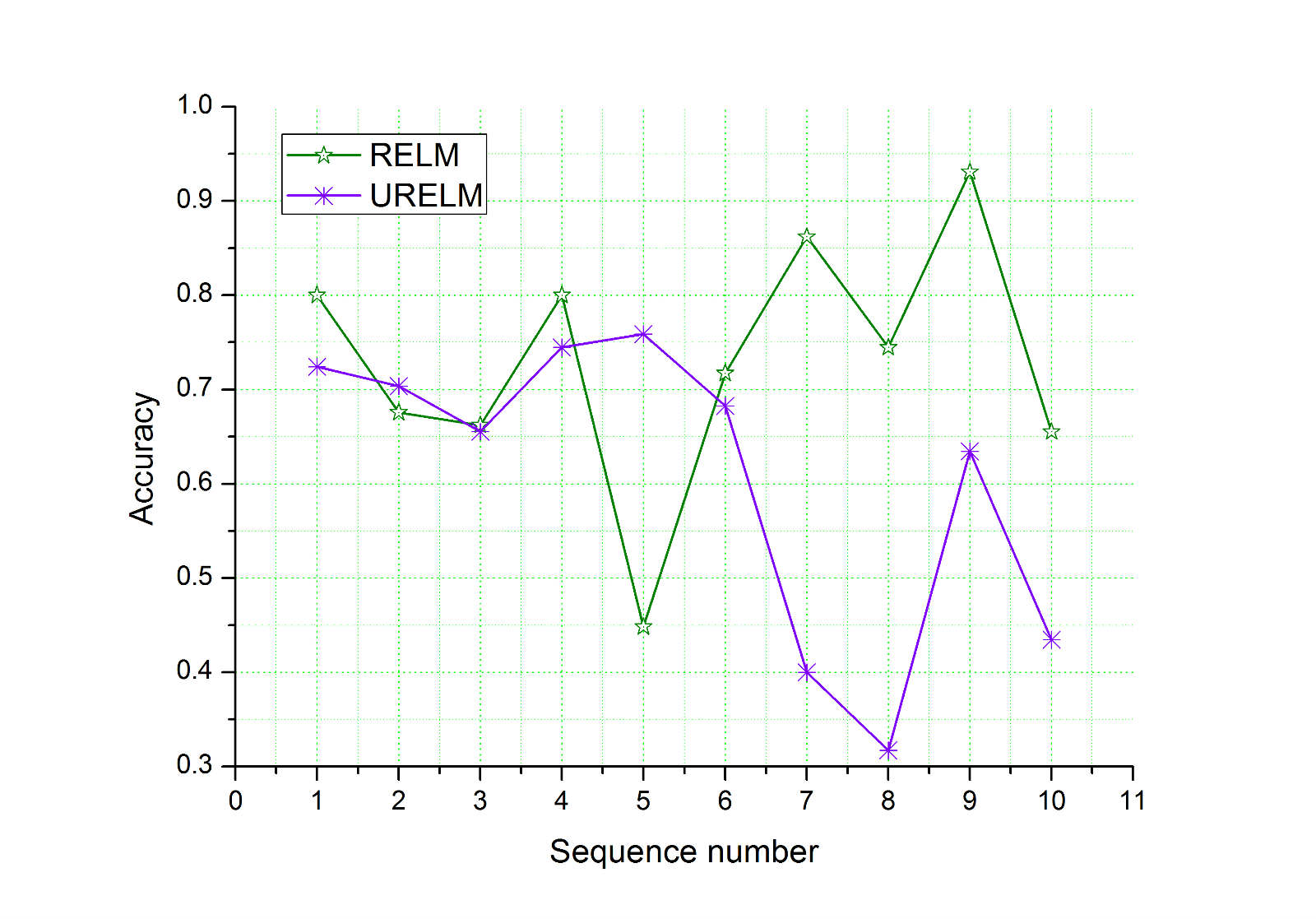}}
    \centerline{3-d votes}
  \end{minipage}
  \hfill
  \begin{minipage}{5cm}
    \centerline{\includegraphics[width=7cm,height=4cm]{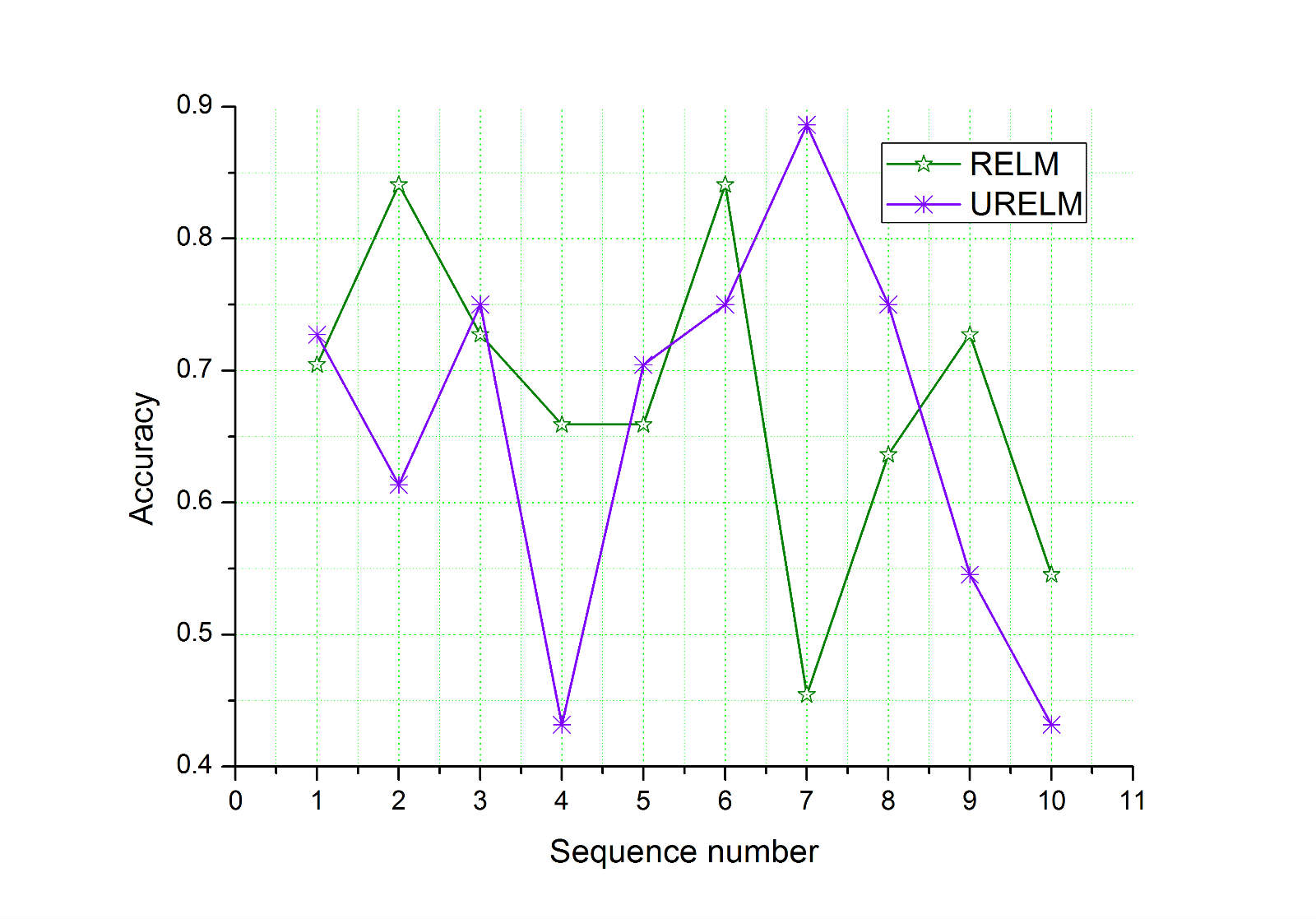}}
    \centerline{3-e Echocardiogram}
  \end{minipage}
  \hfill
  \begin{minipage}{5cm}
    \centerline{\includegraphics[width=7cm,height=4cm]{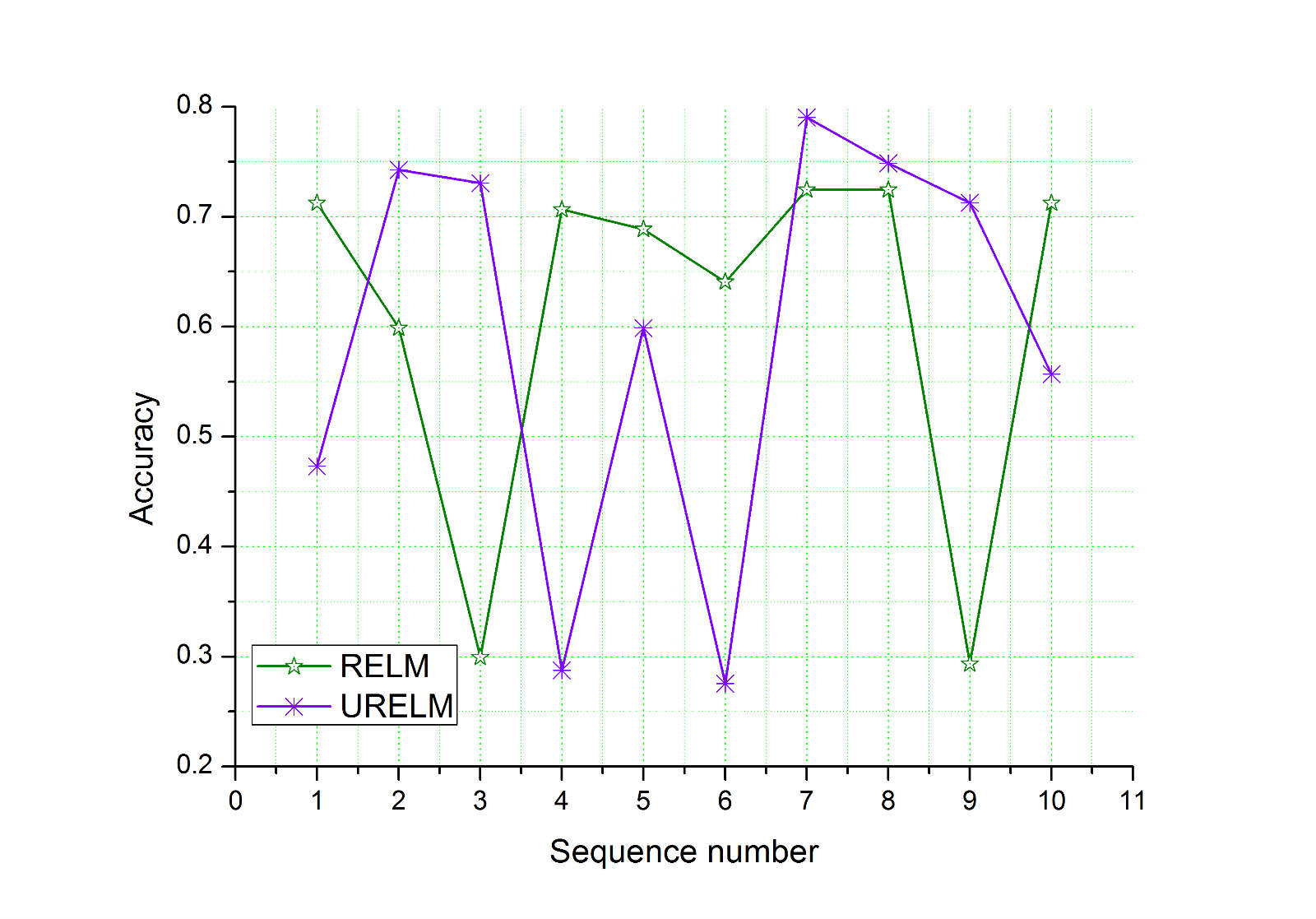}}
    \centerline{3-f Germany}
  \end{minipage}
  \vfill

\begin{minipage}{5cm}
    \centerline{\includegraphics[width=7cm,height=4cm]{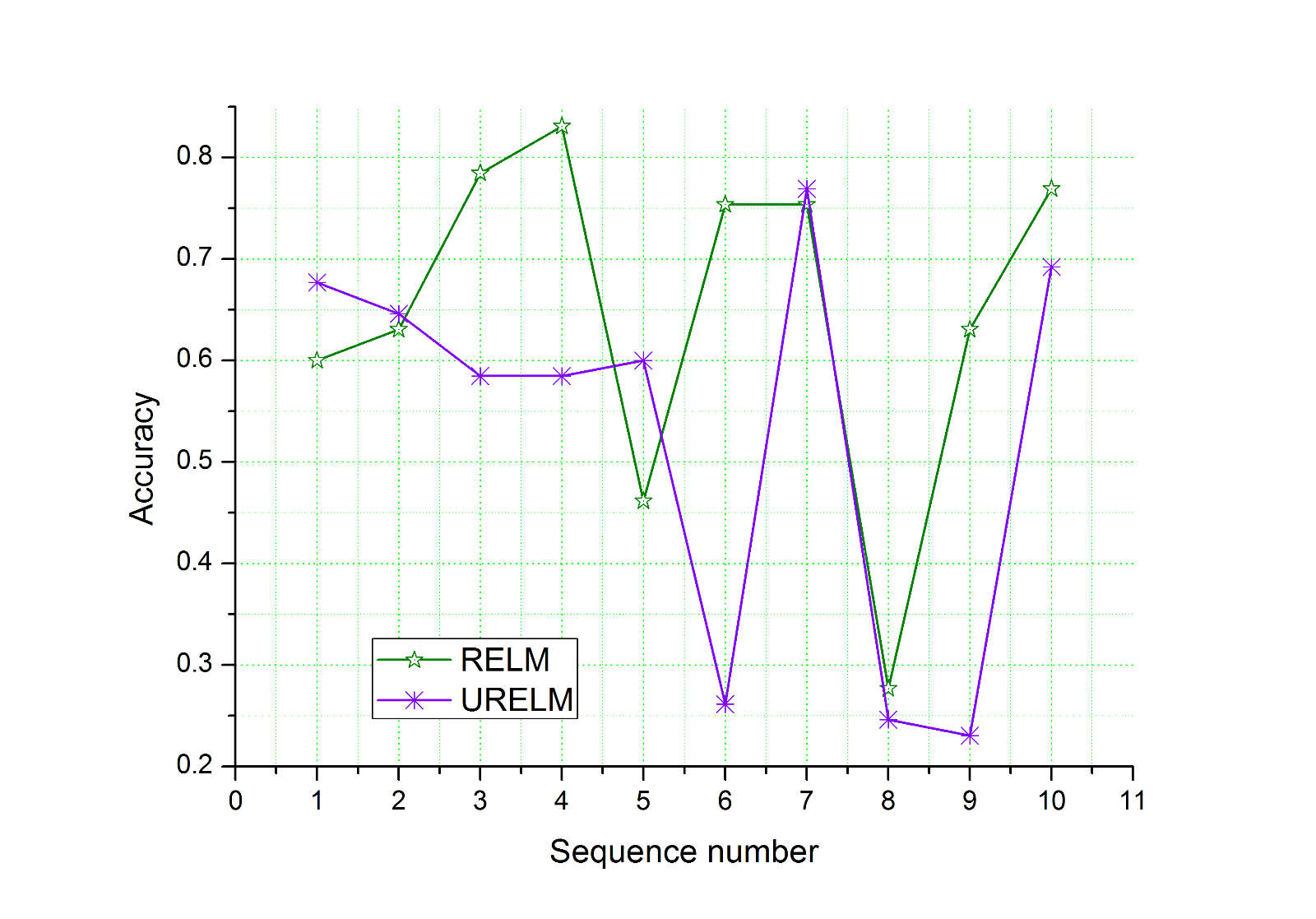}}
    \centerline{3-g parkinsons}
  \end{minipage}
  \hfill
  \begin{minipage}{5cm}
    \centerline{\includegraphics[width=7cm,height=4cm]{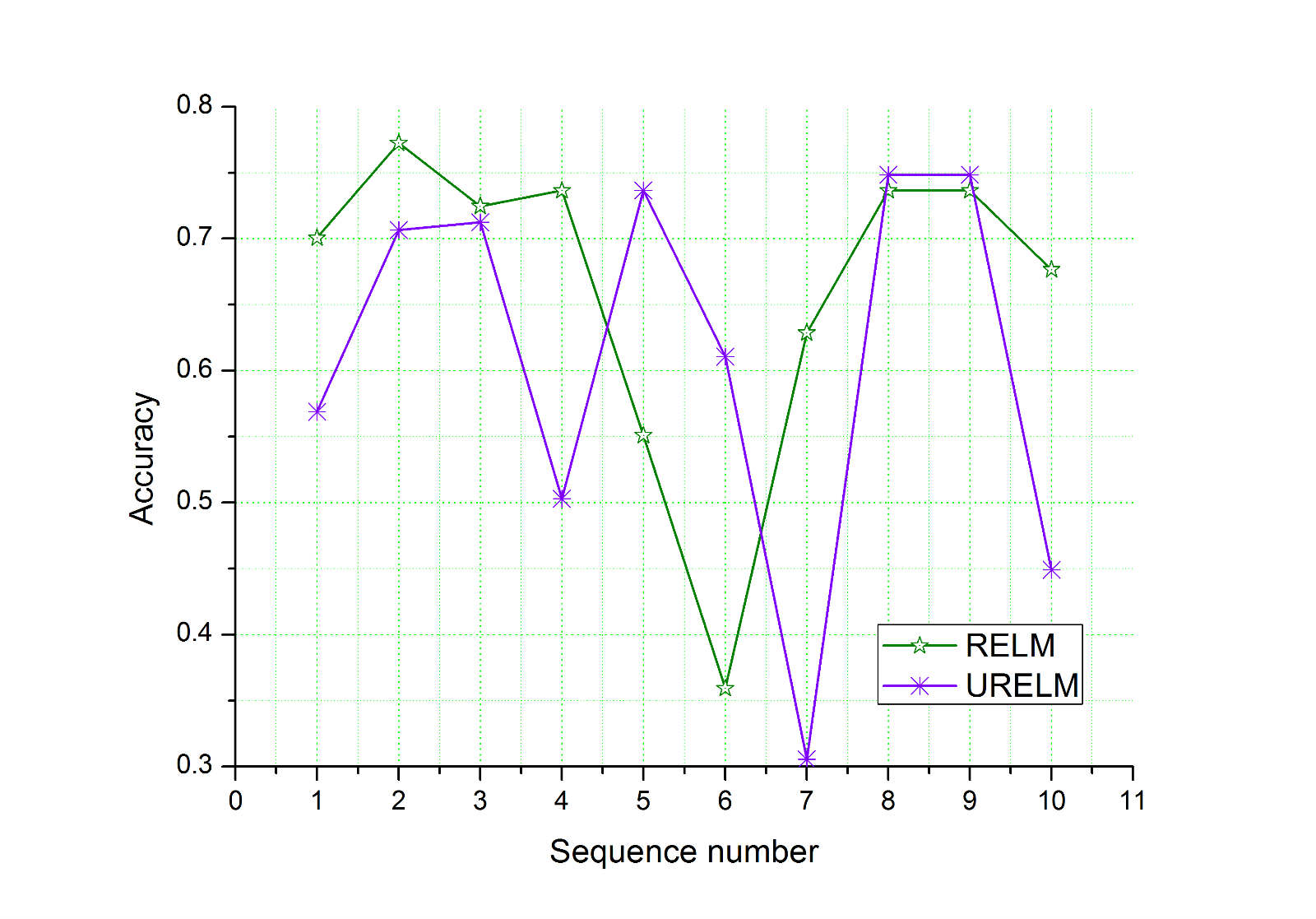}}
    \centerline{3-h yeast}
  \end{minipage}
  \hfill
  \begin{minipage}{5cm}
    \centerline{\includegraphics[width=7cm,height=4cm]{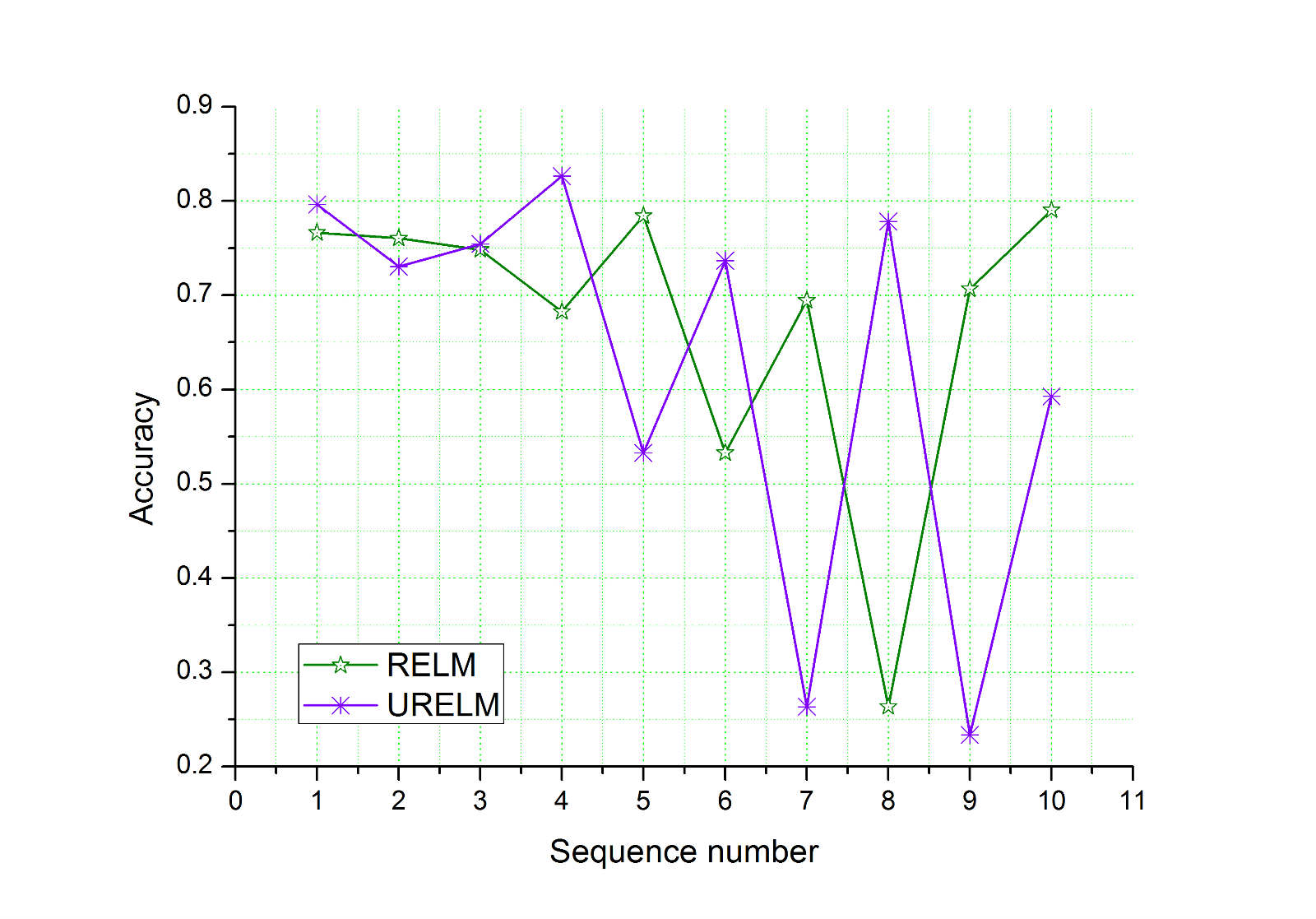}}
    \centerline{3-i vehicle}
  \end{minipage}
  \vfill

\begin{minipage}{5cm}
    \centerline{\includegraphics[width=7cm,height=4cm]{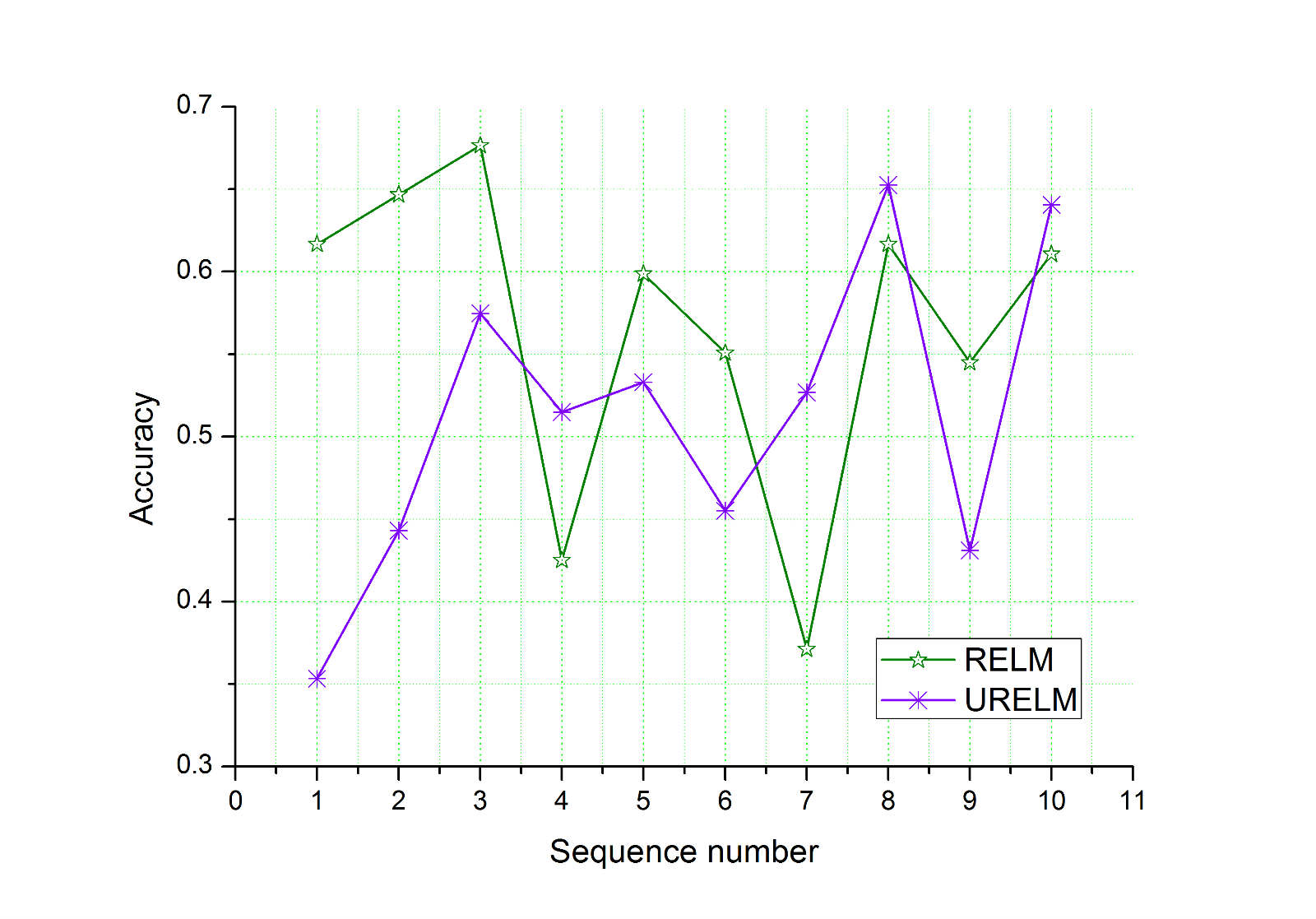}}
    \centerline{3-j pima}
  \end{minipage}
  \hfill
  \begin{minipage}{5cm}
    \centerline{\includegraphics[width=7cm,height=4cm]{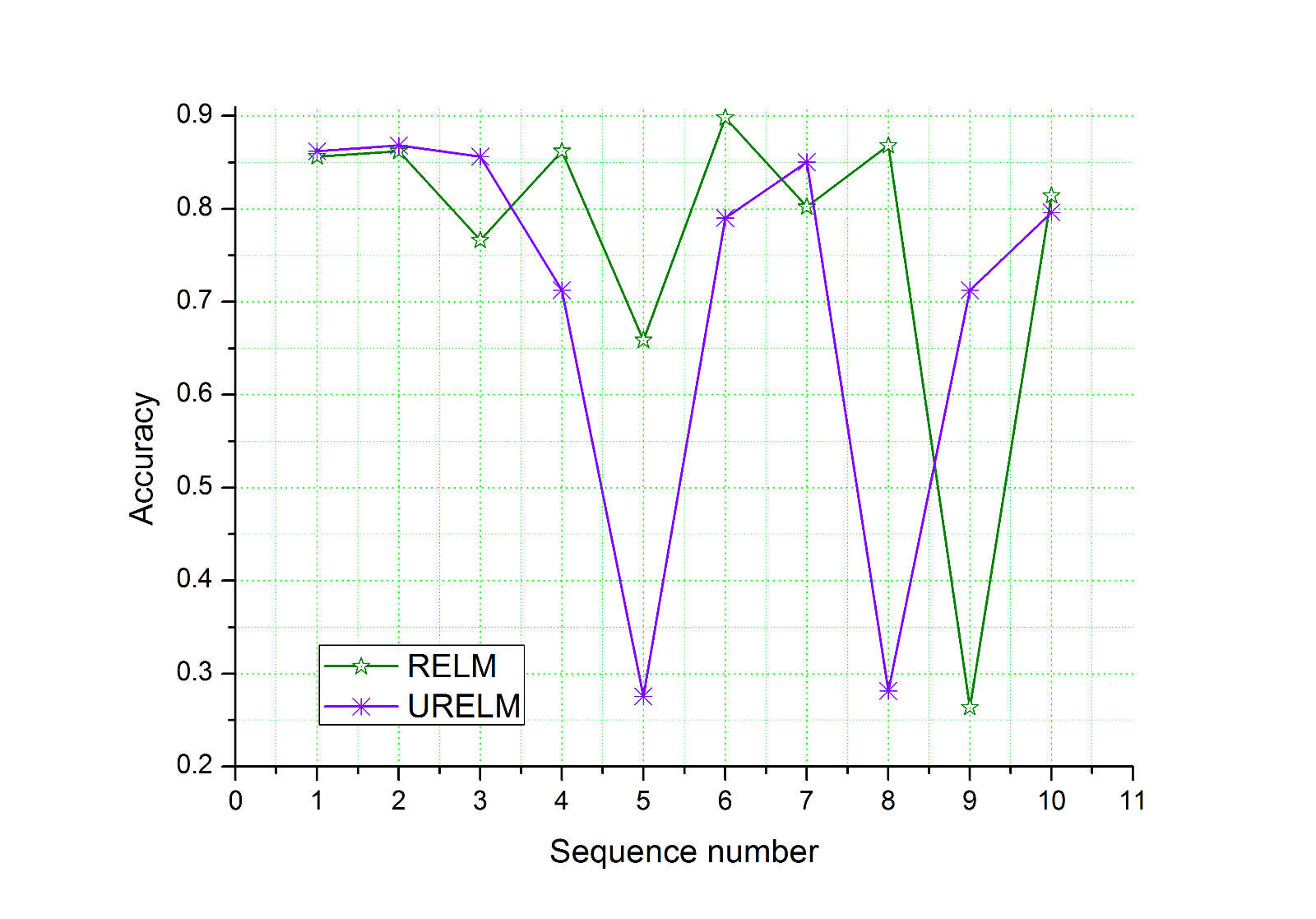}}
    \centerline{3-k segment}
  \end{minipage}
  \hfill
  \begin{minipage}{5cm}
    \centerline{\includegraphics[width=7cm,height=4cm]{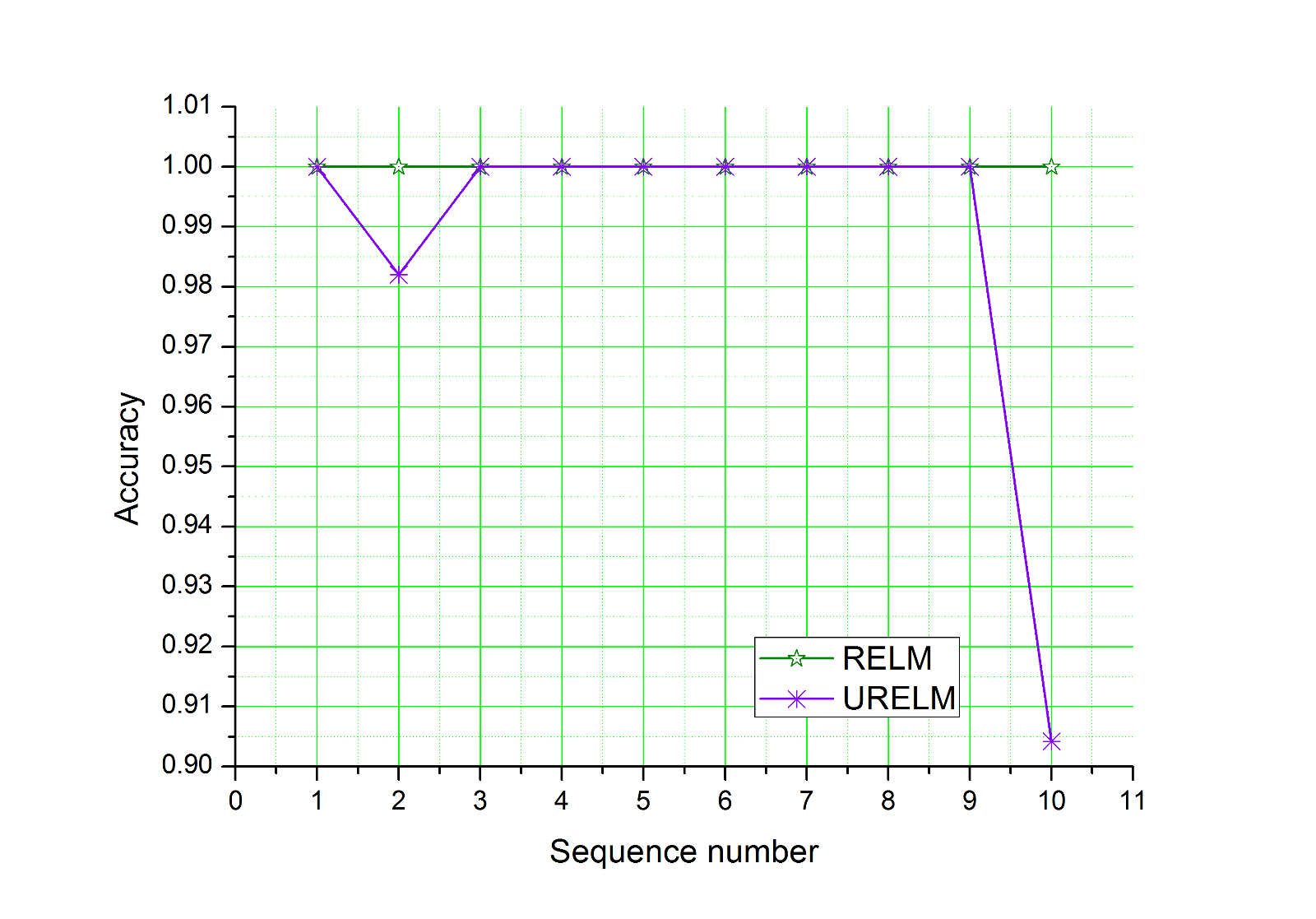}}
    \centerline{3-l STAGGER}
  \end{minipage}
  \vfill

\begin{minipage}{5cm}
    \centerline{\includegraphics[width=7cm,height=4cm]{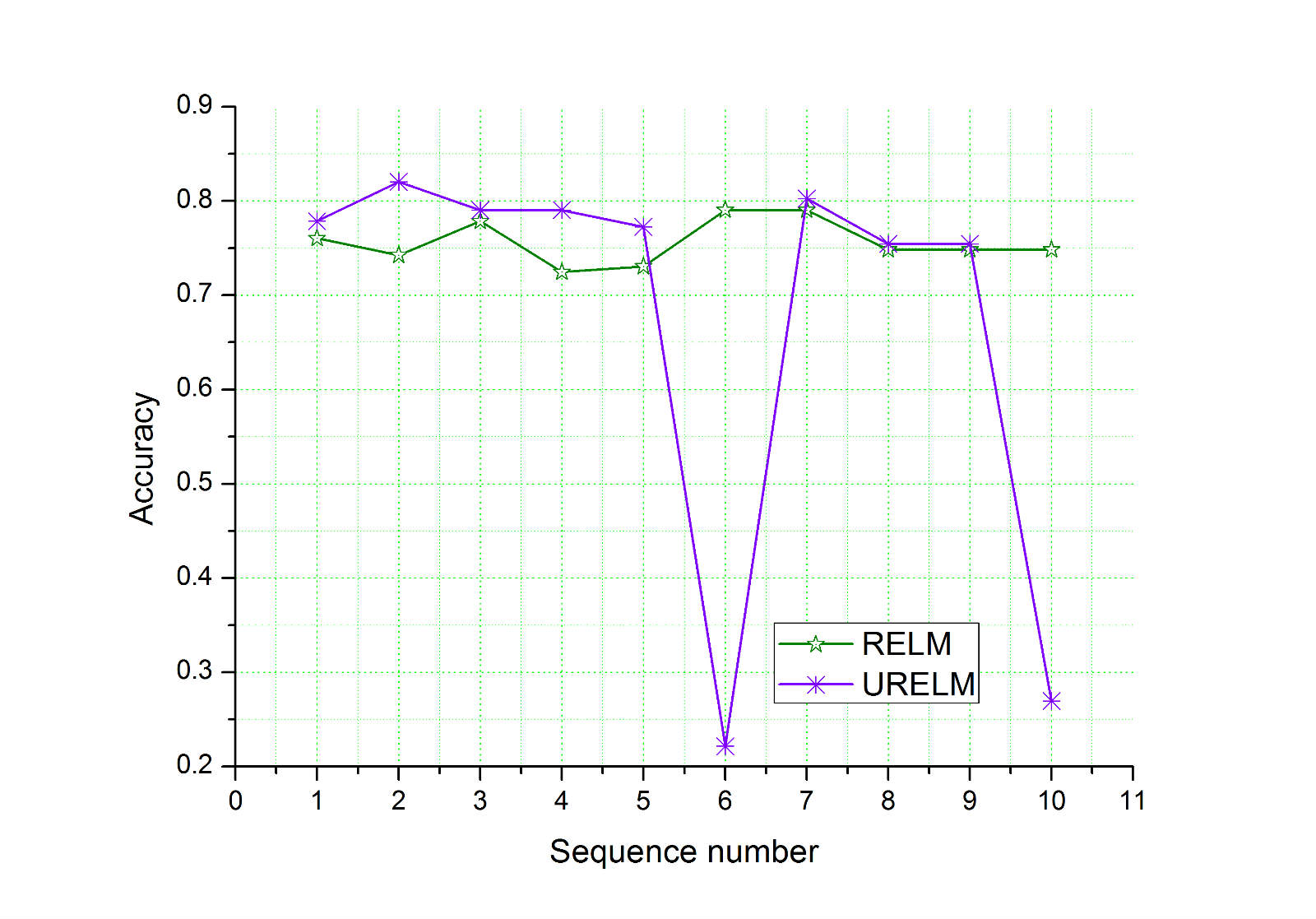}}
    \centerline{3-m adult}
  \end{minipage}
  \hfill
  \begin{minipage}{5cm}
    \centerline{\includegraphics[width=7cm,height=4cm]{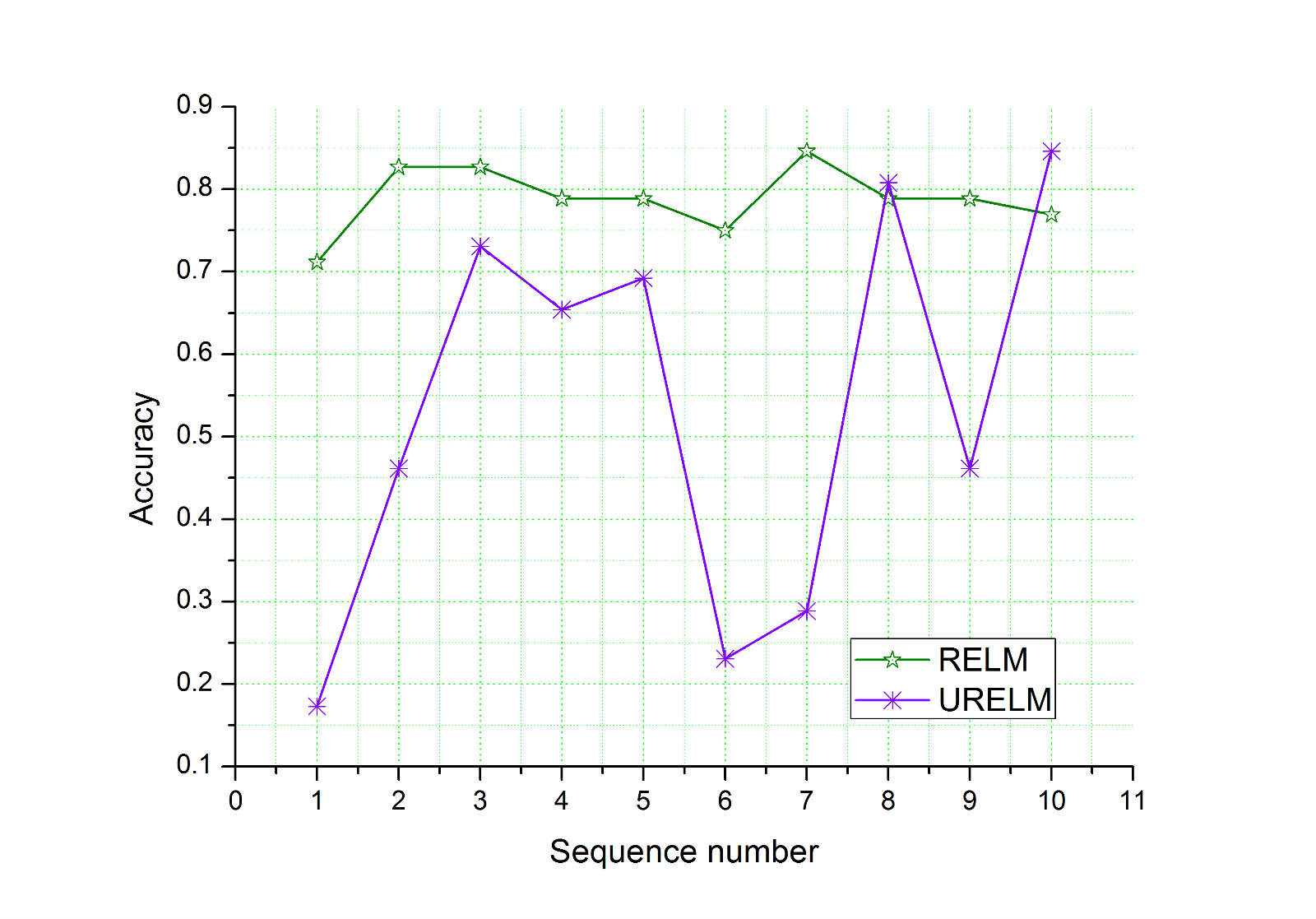}}
    \centerline{3-n Hepatitis}
  \end{minipage}
\centering
\caption{The test results of RELM and URELM on different data sets}
\end{figure}

\begin{table}[h]
  \centering
  \footnotesize
  \caption{The  accuarcies of RELM and URELM on different data sets}
  \begin{tabular}{cccc}\toprule
    Algorithms & RELM & URELM \\\midrule
    Horse & \textbf{0.694$\pm$0.1625} & 0.5460$\pm$0.2375 \\
    glass & \textbf{0.8656$\pm$0.0528} & 0.8095$\pm$0.2398 \\
    haberman & \textbf{0.6228$\pm$0.1476} & 0.5762$\pm$0.1881 \\
    votes & \textbf{0.7297$\pm$0.1336} & 0.6055$\pm$0.1600 \\
    Echocardiogram & \textbf{0.6796$\pm$0.1195} & 0.6591$\pm$0.1496 \\
    Germany & \textbf{0.6102$\pm$0.1702} & 0.5916$\pm$0.1911 \\
    parkinsons & \textbf{0.6492$\pm$0.1714} & 0.5292$\pm$0.2032 \\
    yeast & \textbf{0.6623$\pm$0.1245} & 0.6089$\pm$0.1513 \\
    vehicle & \textbf{0.6731$\pm$0.1624} & 0.6246$\pm$0.2179 \\
    pima & \textbf{0.5659$\pm$0.0974} & 0.5125$\pm$0.0947 \\
    segment & \textbf{0.7653$\pm$0.1893} & 0.7006$\pm$0.2297 \\
    STAGGER & \textbf{1.0000$\pm$0.0000} & 0.9886$\pm$0.0302 \\
    adult & \textbf{0.7563$\pm$0.0233} & 0.6755$\pm$0.2278 \\
    Hepatitis & \textbf{0.7885$\pm$0.0395} & 0.5346$\pm$0.2457 \\\bottomrule
  \end{tabular}
\end{table}

\begin{table}[h]
  \centering
  \footnotesize
  \caption{The  Reduction efficiency of RELM algorithm on the experimental data sets}
  \begin{tabular}{ccccccc}\toprule
    Data sets & Before reduction & After reduction & Reduced dimensions & Reduced dimensions/Befor reduction\\\midrule
    Horse &26 &3 &23 &0.8846 \\
    glass &9 &4 &5 &0.5556 \\
    haberman &3 &2 &1 &0.3333 \\
    votes &16 &9 &7 &0.4375 \\
    Echocardiogram &12 &9 &3 &0.2500 \\
    Germany &24 &7 &17 &0.7083 \\
    parkinsons &22 &7 &15 &0.6818 \\
    yeast &8 &8 &0 &0.0000 \\
    vehicle &18 &13 &5 &0.2778 \\
    pima &8 &8 &0 &0.0000 \\
    segment &19 &11 &8 &0.4211 \\
    STAGGER &3 &2 &1 &0.3333 \\
    adult &13 &12 &1 &0.0769 \\
    Hepatitis &19 &19 &19 &1.0000 \\\bottomrule
  \end{tabular}
\end{table}

\subsection{The impact of data dimensions on time overheard}

In order to test the impact of data dimensions on time overheard, we choose \emph{Horse, biodeg, lungcancer, Germany, parkinsons, vehicle, segement, Hypelane} and \emph{Waveform} as experimental data sets; the activation function is \emph{radbas}; the number of neurons in hidden layer \emph{L} is 650 and \emph{C} is 1000. BP algorithm \cite{Han2001Data} with rough set reduction method (remarked as RS+BP) is as comparison algorithm. The time overheard is showed as Fig.4-12.

\begin{figure}[H]
\begin{minipage}[t]{0.3\linewidth}
\centering
\includegraphics[height=5cm,width=6cm]{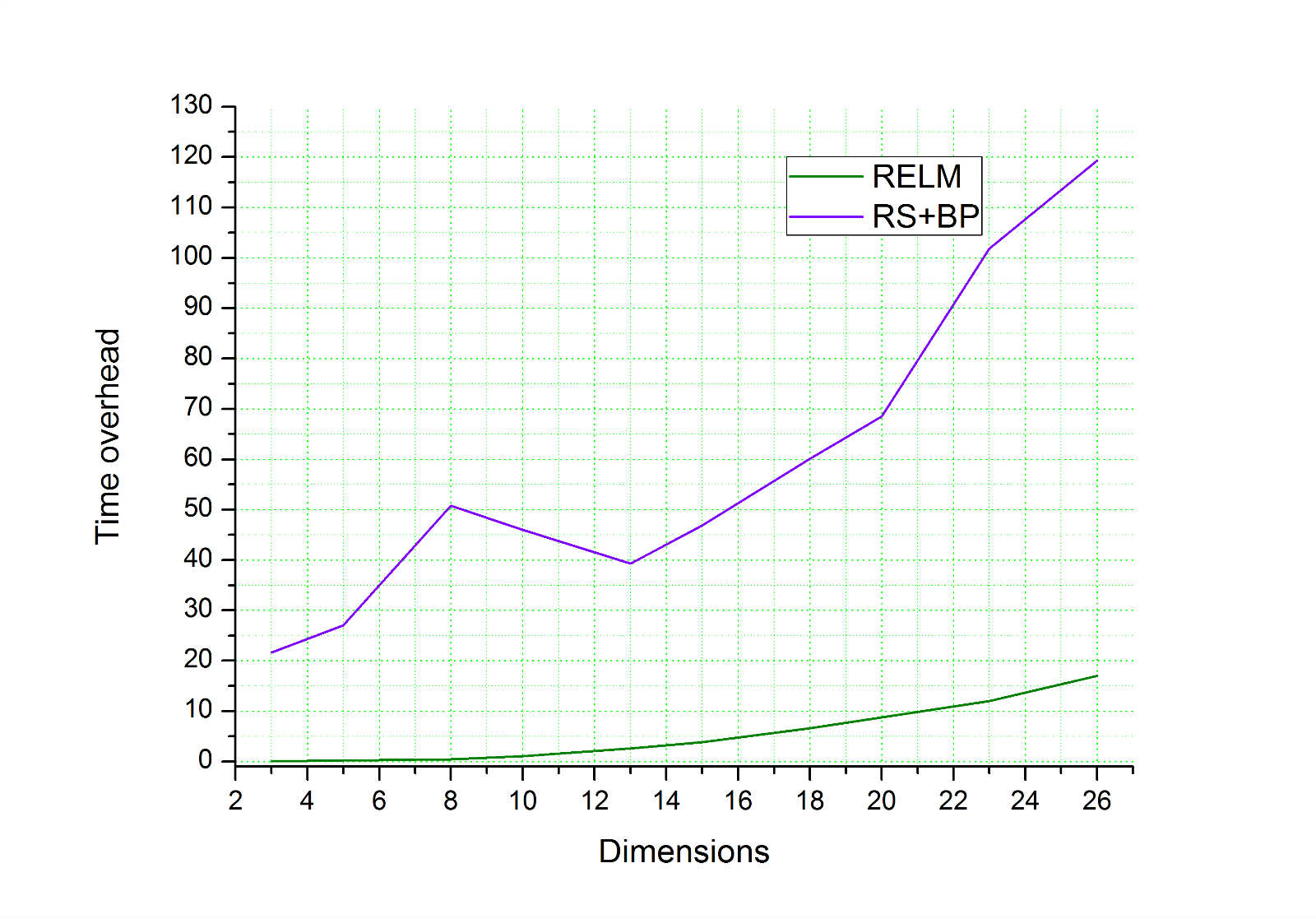}
\caption{Horse}
\end{minipage}%
\hfill
\begin{minipage}[t]{0.3\linewidth}
\centering
\includegraphics[height=5cm,width=6cm]{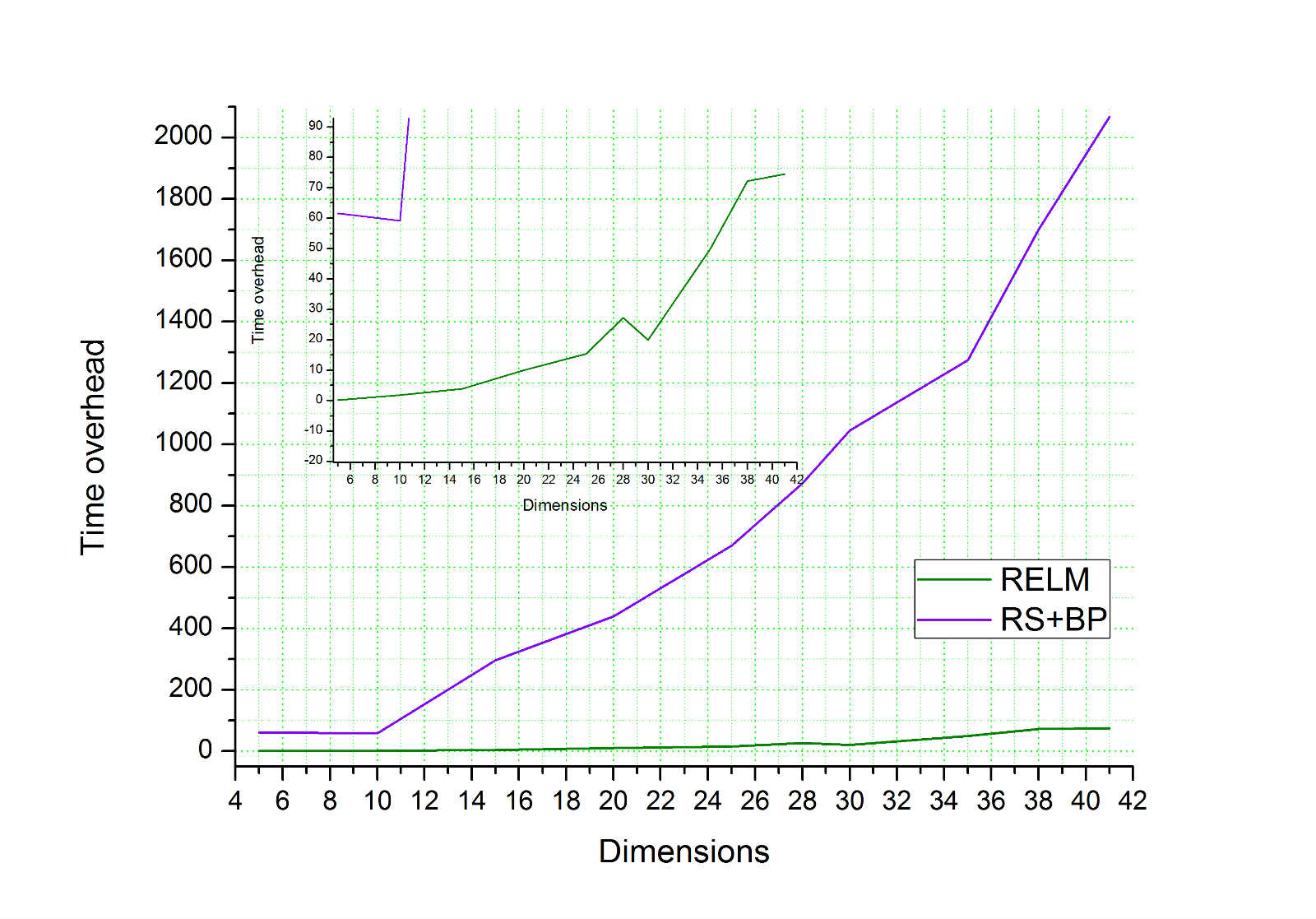}
\caption{biodeg}
\end{minipage}
\hfill
\begin{minipage}[t]{0.3\linewidth}
\centering
\includegraphics[height=5cm,width=6cm]{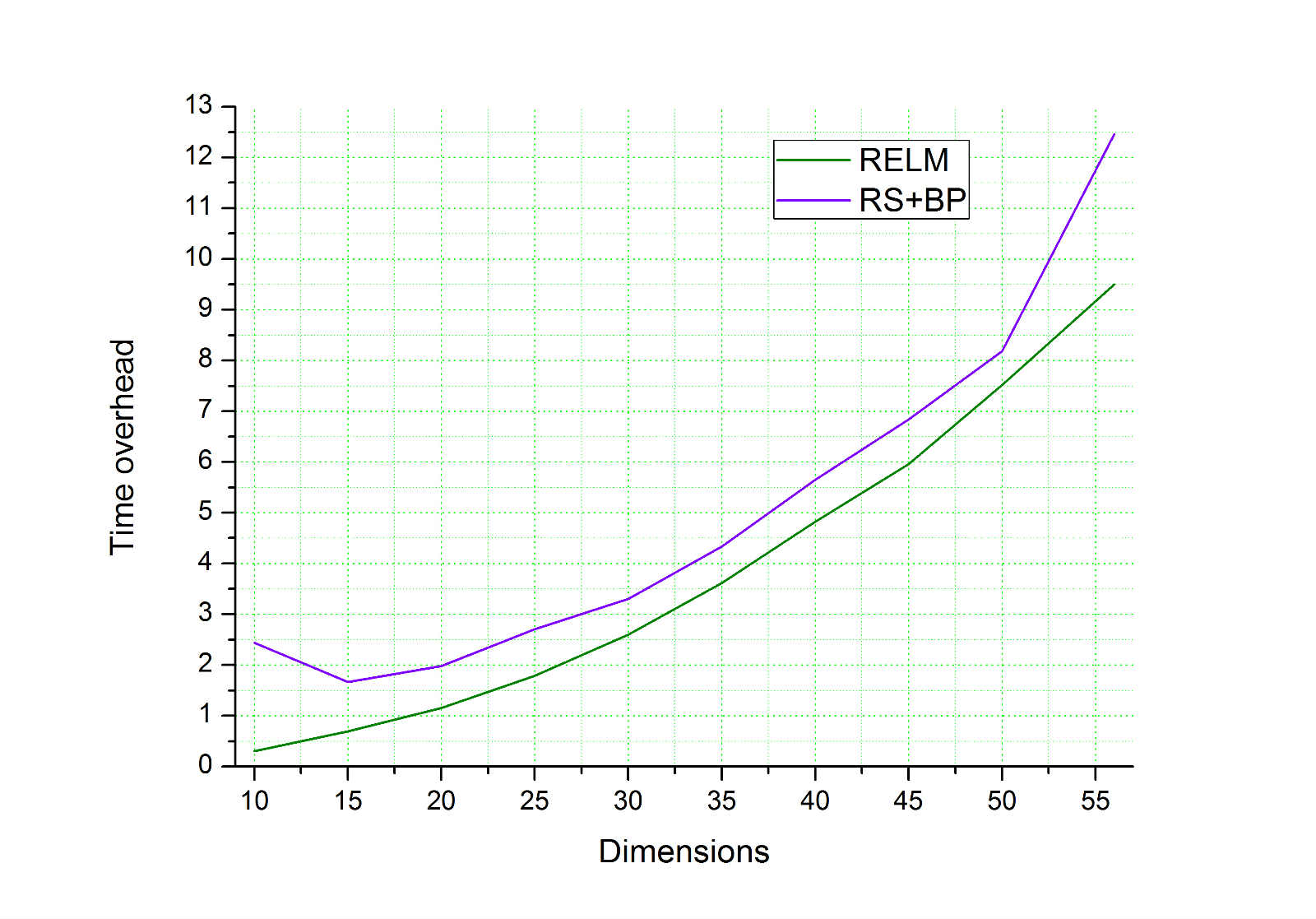}
\caption{lungcancer}
\end{minipage}
\end{figure}

\begin{figure}[H]
\begin{minipage}[t]{0.3\linewidth}
\centering
\includegraphics[height=6cm,width=6cm]{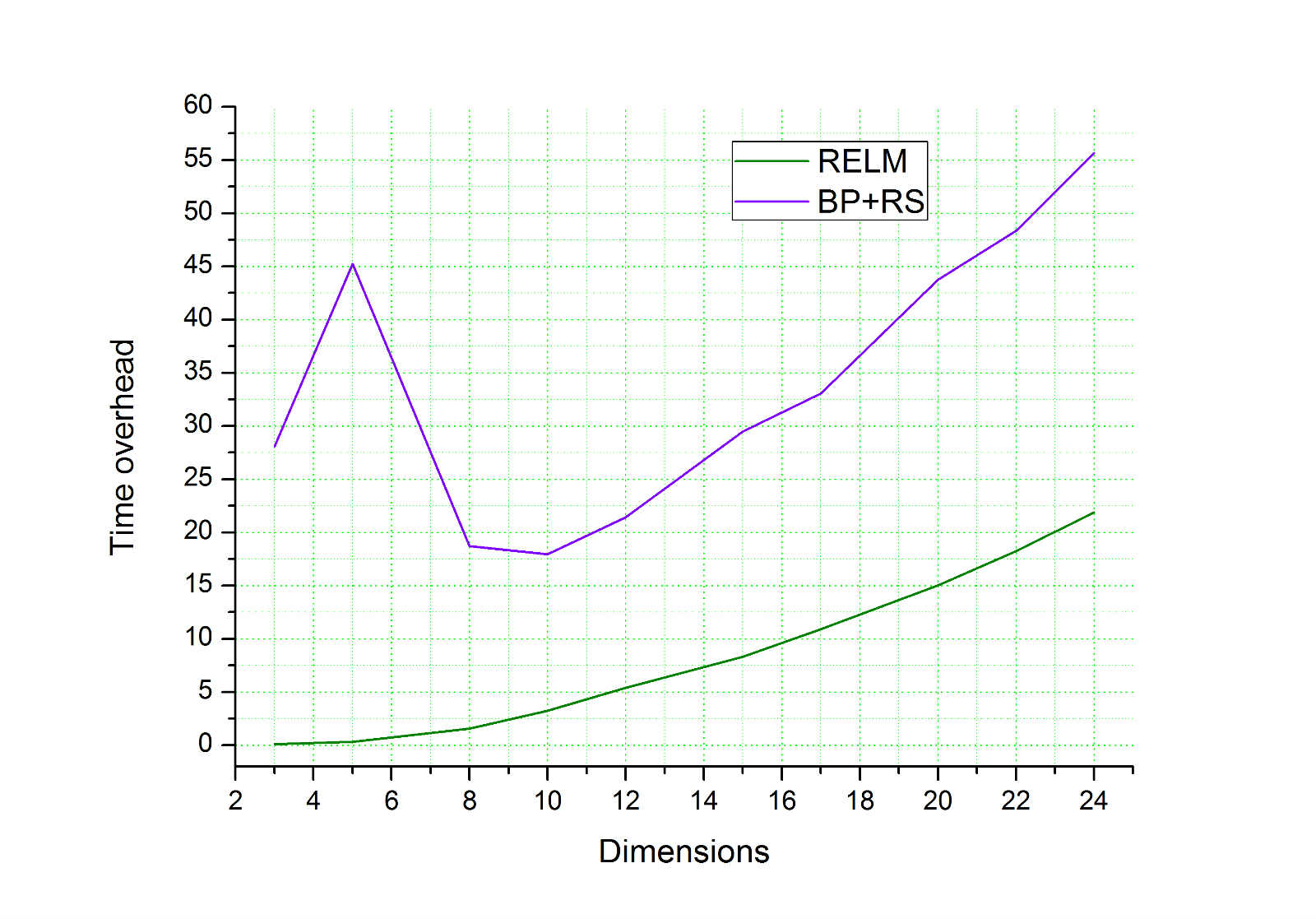}
\caption{Germany}
\end{minipage}%
\hfill
\begin{minipage}[t]{0.3\linewidth}
\centering
\includegraphics[height=6cm,width=6cm]{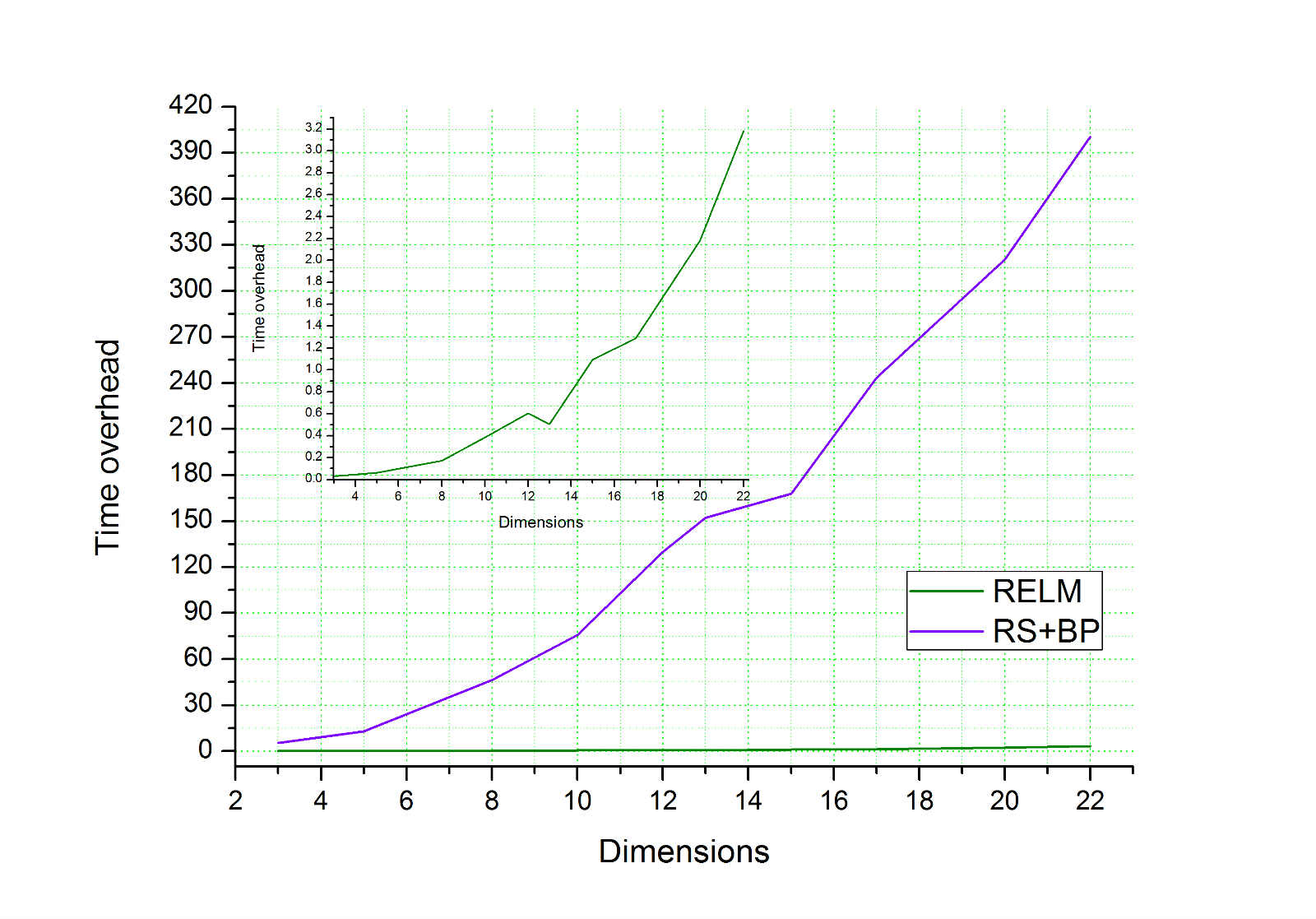}
\caption{parkinsons}
\end{minipage}
\hfill
\begin{minipage}[t]{0.3\linewidth}
\centering
\includegraphics[height=6cm,width=6cm]{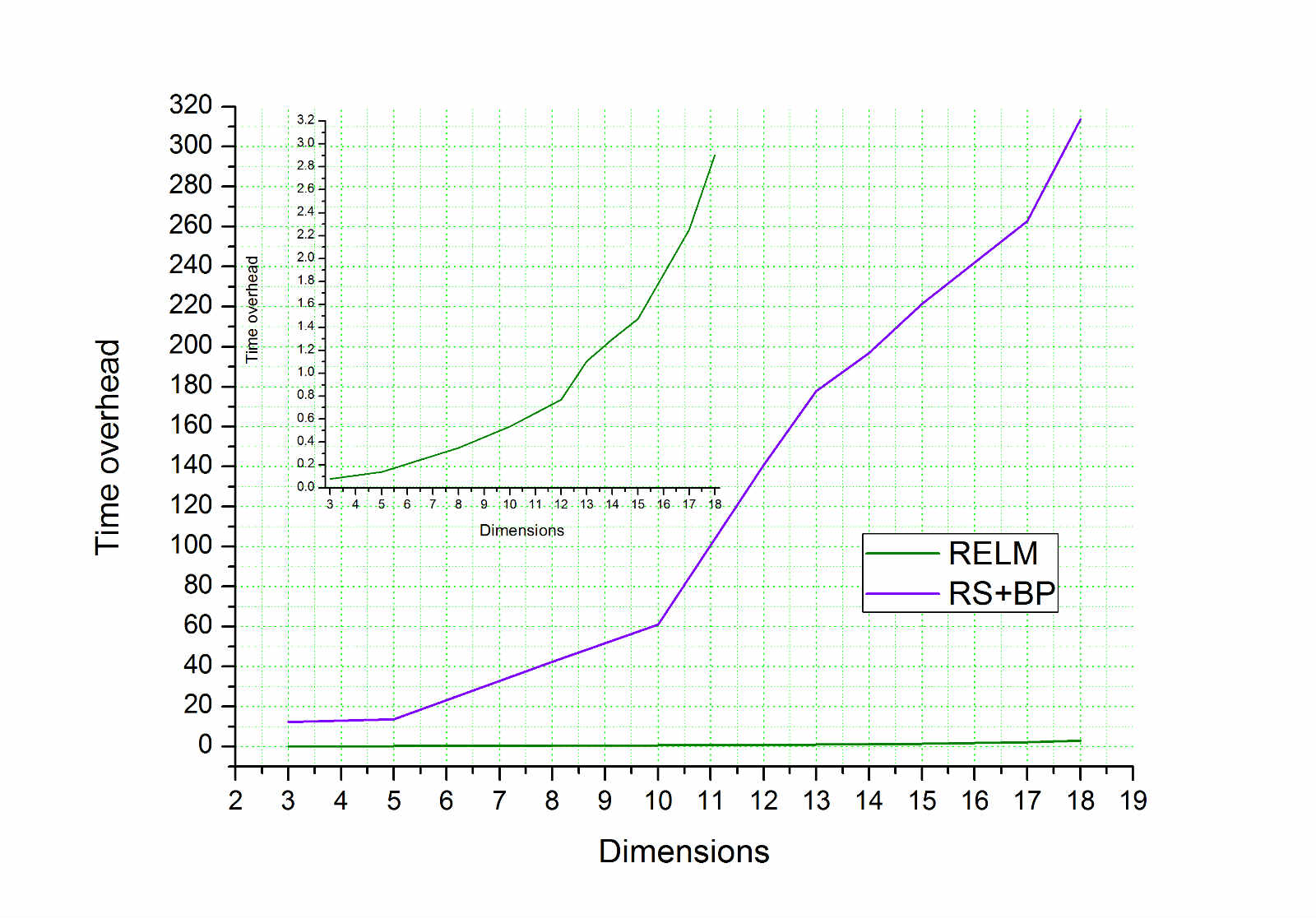}
\caption{vehicle}
\end{minipage}
\end{figure}

\begin{figure}[H]
\begin{minipage}[t]{0.3\linewidth}
\centering
\includegraphics[height=6cm,width=6cm]{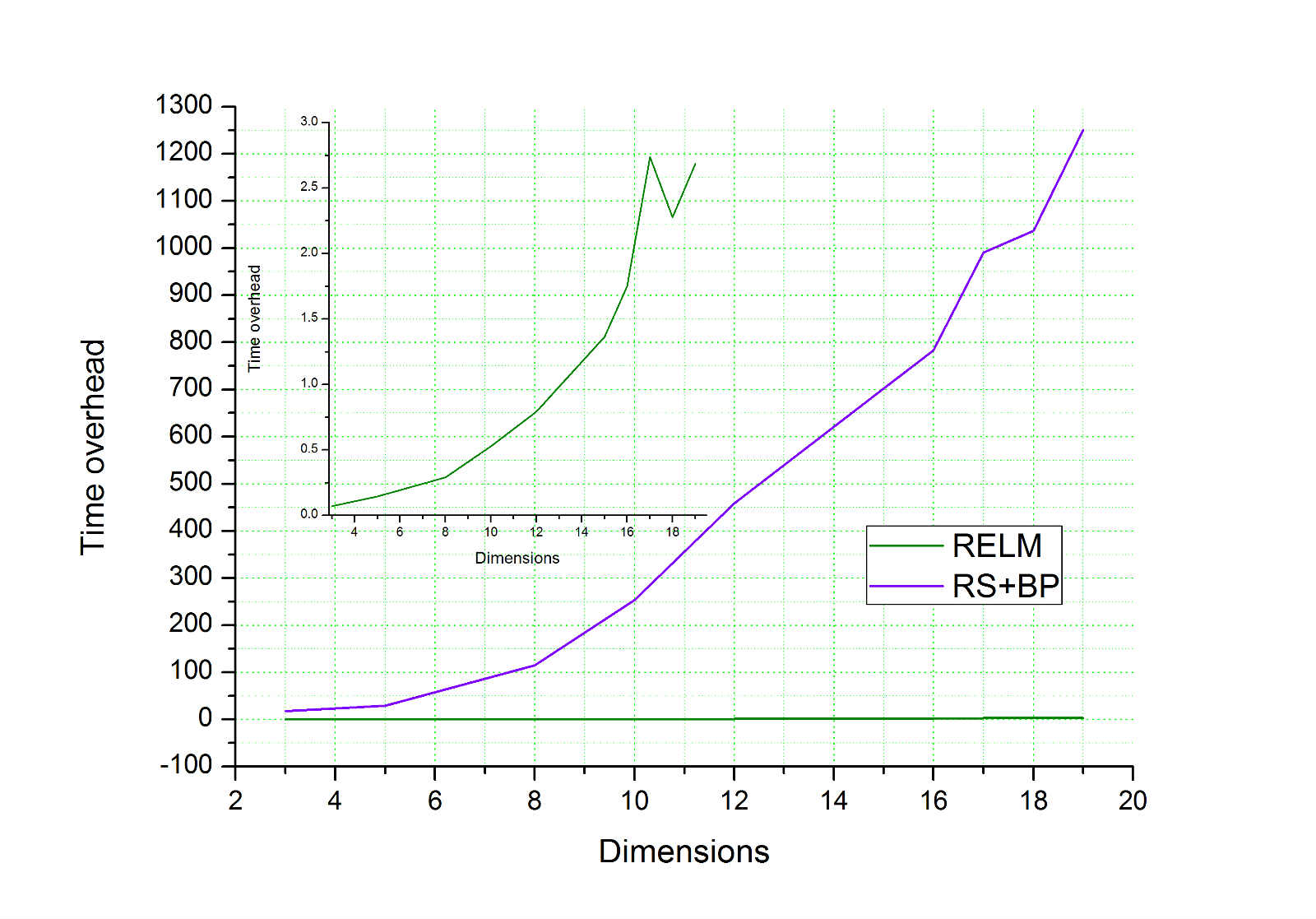}
\caption{segment}
\end{minipage}%
\hfill
\begin{minipage}[t]{0.3\linewidth}
\centering
\includegraphics[height=6cm,width=6cm]{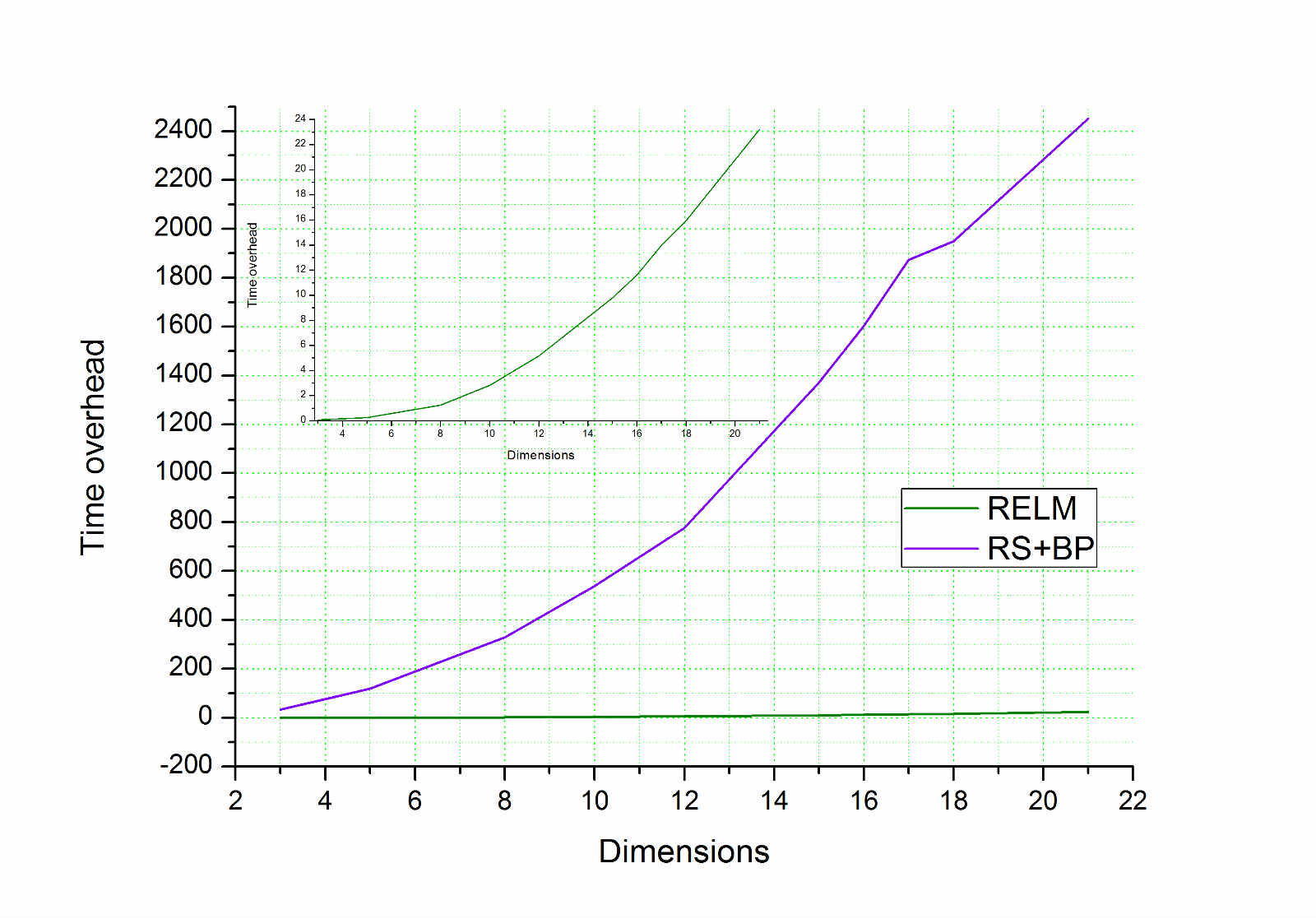}
\caption{Waveform}
\end{minipage}
\hfill
\begin{minipage}[t]{0.3\linewidth}
\centering
\includegraphics[height=6cm,width=6cm]{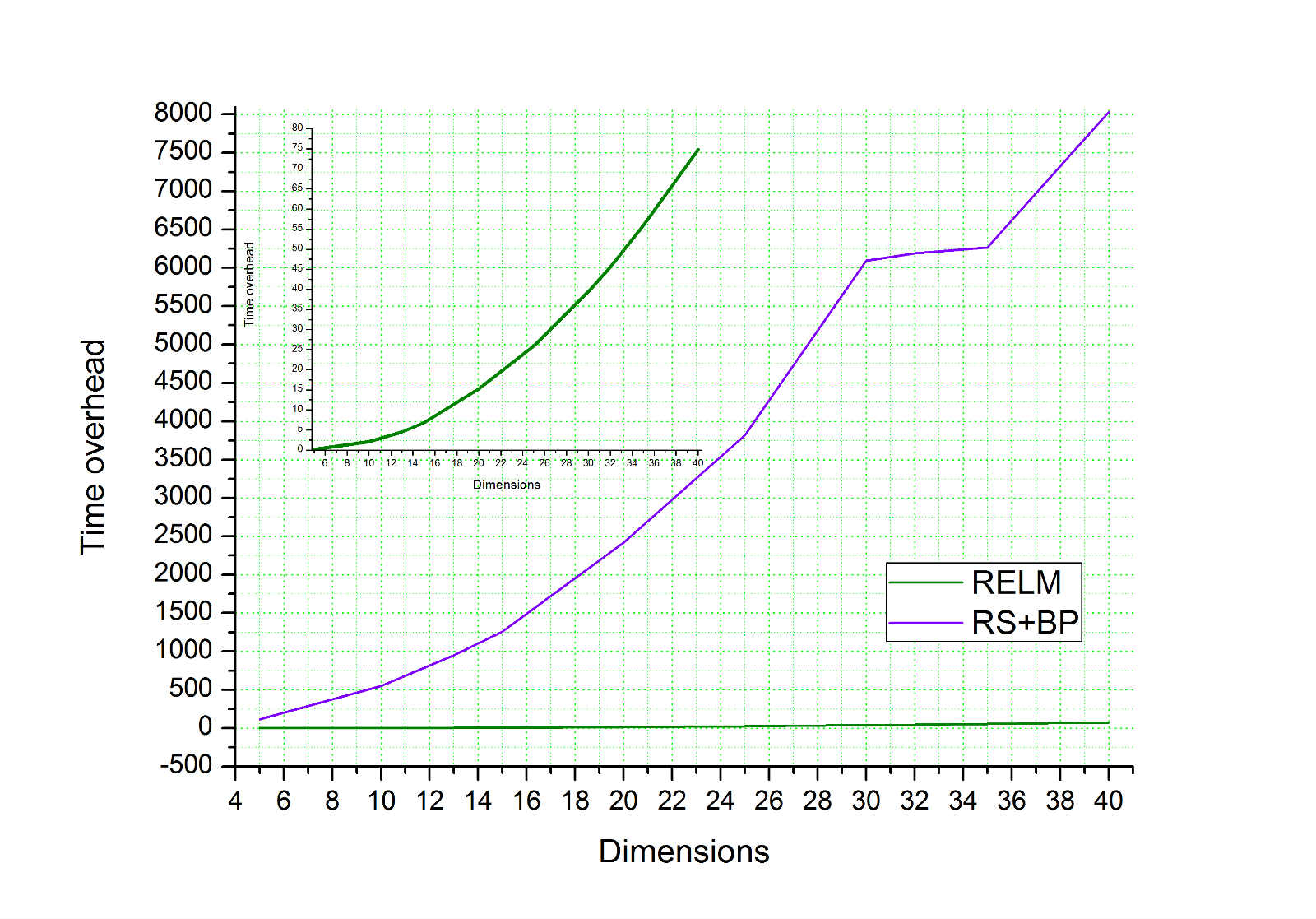}
\caption{Hyperplane}
\end{minipage}
\end{figure}
Fig.4-Fig.12 are the time overhead of RELM and RS+BP. The subgraph in each graph is the local graph of the change trend of RELM's time overhead. From Fig.4-Fig.12, it can be seen that the time overhead of RELM and RS+BP increases with the increase of data dimensions. RS+BP is a very time-consuming algorithm, and the time overhead of RS+BP is much more than that of RELM. Because RELM has utilized the ELM mechanism, the cost of RELM Keeps at a lower level although the time expenditure is also showing a significant increasing trend. In a word, the time complexity of RELM is less sensitive to data dimension than RS+BP.
\section{Conclusion and future work}
In this paper, we proposed a new extreme learning machine algorithm with rough set method called RELM. RELM utilizes the data division result of rough set to train upper approximation neurons and lower approximation neurons; and the output weights can be analytically determined. The final classification result  is decided by the two kinds of neurons. In addition, the attribute reduction method is introduced to remove redundant attributes. The experimental results showed that RELM is an effective algorithm. However, in some experiments, it can be found the performance of RELM seems unstable; the variances of accuracies is somewhat large on some data sets. From Fig.4-Fig.12, it can be seen that the time cost of RELM almost increases exponentially. So how to improve the performance's stability of RELM and further reduce the time complexity will be research directions in our future work.

\section*{Acknowledgments}
This work was supported by National Natural Science Fund of China (Nos.61672130, 61602082, 61370200), the Open Program of State Key Laboratory of Software Architecture (No. SKLSAOP1701), China Postdoctoral Science Foundation (No.2015M581331), Foundation of LiaoNing Educational Committee (Nos.201602151) and MOE Research Center for Online Education (2016YB121).

\section*{References}

\small{
\bibliographystyle{elsarticle-num}
\bibliography{bibfile}
}
\end{document}